\documentclass[10pt,twocolumn,letterpaper]{article}

\usepackage{iccv_arxiv}              
\usepackage{tcolorbox}

\def\eg{{\em e.g.}}

\def\etc{{\em etc.}}
\newcommand{\cmtt}[1]

\DeclareRobustCommand{\name}{RoboTron-Drive}

\usepackage{graphicx}
\usepackage{booktabs}
\usepackage{graphicx}
\usepackage{amsmath}
\usepackage{amssymb}
\usepackage{booktabs}
\usepackage{multirow}
\usepackage{tabularx}
\usepackage{enumitem}
\usepackage{gensymb}
\usepackage{colortbl}
\usepackage{caption}
\usepackage{makecell}
\usepackage{float} 
\usepackage{placeins}
\usepackage{wrapfig}
\usepackage{amssymb}
\usepackage{caption} 
\usepackage{pifont}
\definecolor{mygray}{gray}{.95}
\definecolor{mycyan}{cmyk}{.1,0,0,0}
\definecolor{darkgreen}{rgb}{0.0, 0.5, 0.0}
\newcommand{\cmark}{\textcolor{darkgreen}{\ding{51}}}
\newcommand{\xmark}{\textcolor{red}{\ding{55}}}
\usepackage[dvipsnames]{xcolor}

\newcounter{mysection} 

\newcommand{\mysection}[1]{
  \stepcounter{mysection} 
  \section*{\Alph{mysection}.\ #1} 
  \addcontentsline{toc}{section}{\Alph{mysection}.\ #1} 
  \renewcommand{\thetable}{S\arabic{table}}
  \renewcommand{\thefigure}{S\arabic{figure}}
  \setcounter{table}{0} 
  \setcounter{figure}{0} 
}

\newcommand{\cmarkg}{\ding{51}}%
\newcommand{\xmarkg}{\textcolor{lightgray}{\ding{55}}}%

\newcommand\blfootnote[1]{%
\begingroup
\renewcommand\thefootnote{}\footnote{#1}%
\addtocounter{footnote}{-1}%
\endgroup
}
\definecolor{iccvblue}{rgb}{0.21,0.49,0.74}
\usepackage[pagebackref,breaklinks,colorlinks,allcolors=iccvblue]{hyperref}

\def\paperID{9595} 
\def\confName{ICCV}
\def\confYear{2025}

\title{\textit{RoboTron-Drive}: All-in-One Large Multimodal Model for Autonomous Driving}
\makeatletter
\newcommand{\printfnsymbol}[1]{%
  \textsuperscript{\@fnsymbol{#1}}%
}
\makeatother

\author{Zhijian Huang$^{1,\dagger}$ \quad Chengjian Feng$^{2,\dagger}$  \quad Feng Yan$^{2}$  \quad Baihui Xiao$^{2}$ \quad Zequn Jie$^{2}$ \\  Yujie Zhong$^{2}$ \quad Xiaodan Liang$^{1,\ddagger}$ \quad Lin Ma$^{2,\ddagger}$
\\
{ $^{1}$Shenzhen Campus of Sun Yat-sen University \quad
 $^{2}$Meituan }\\
\url{https://zhijian11.github.io/RoboTron-Drive}
}

\begin{document}



\twocolumn[{%
\renewcommand\twocolumn[1][]{#1}%
\maketitle

\vspace{-1.3cm}
\begin{center}
   \begin{minipage}{0.48\linewidth}
       \centering
       \includegraphics[width=\linewidth]{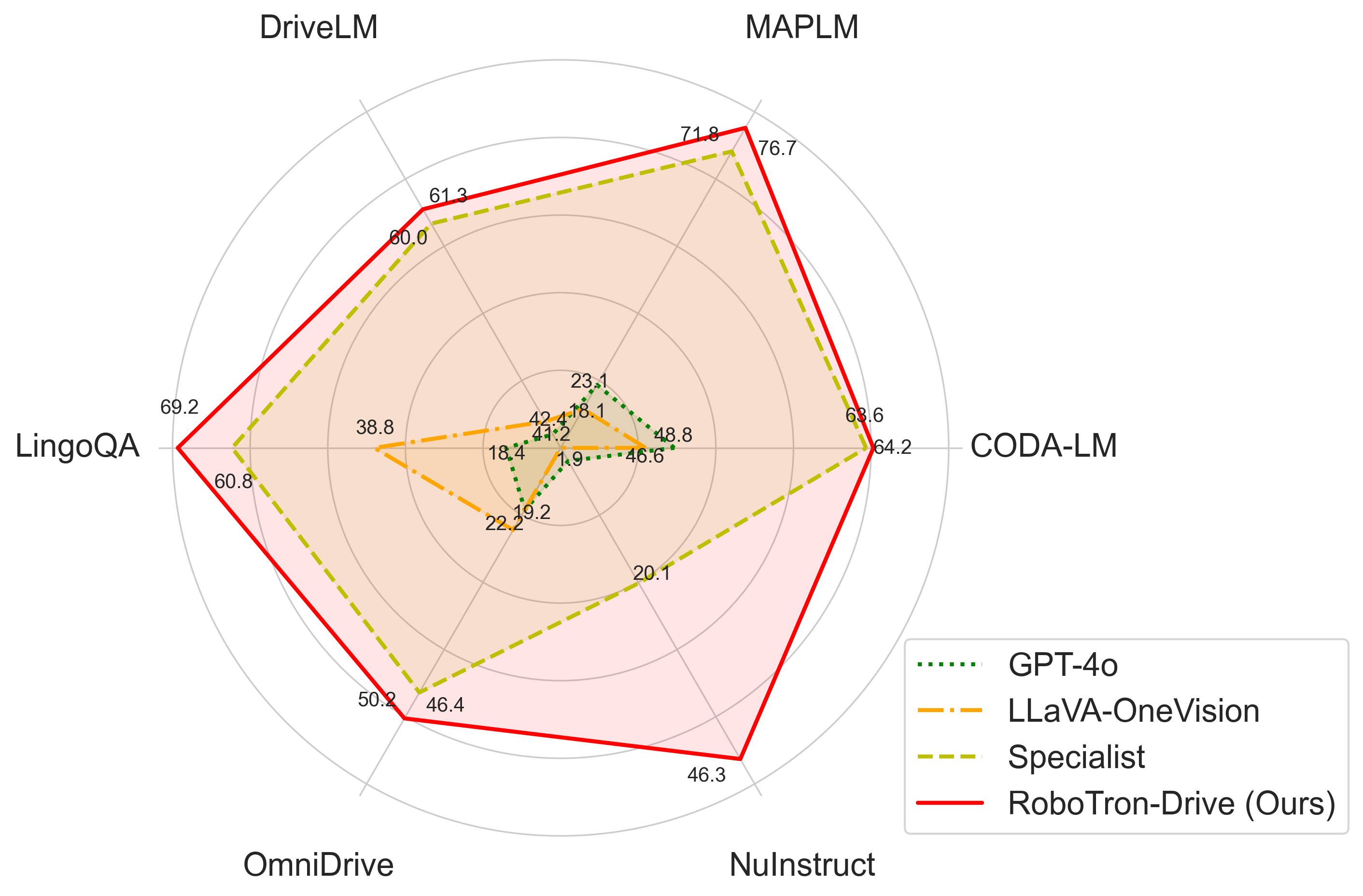}
   \end{minipage}\hfill
   \begin{minipage}{0.48\linewidth}
       \centering
       \includegraphics[width=\linewidth]{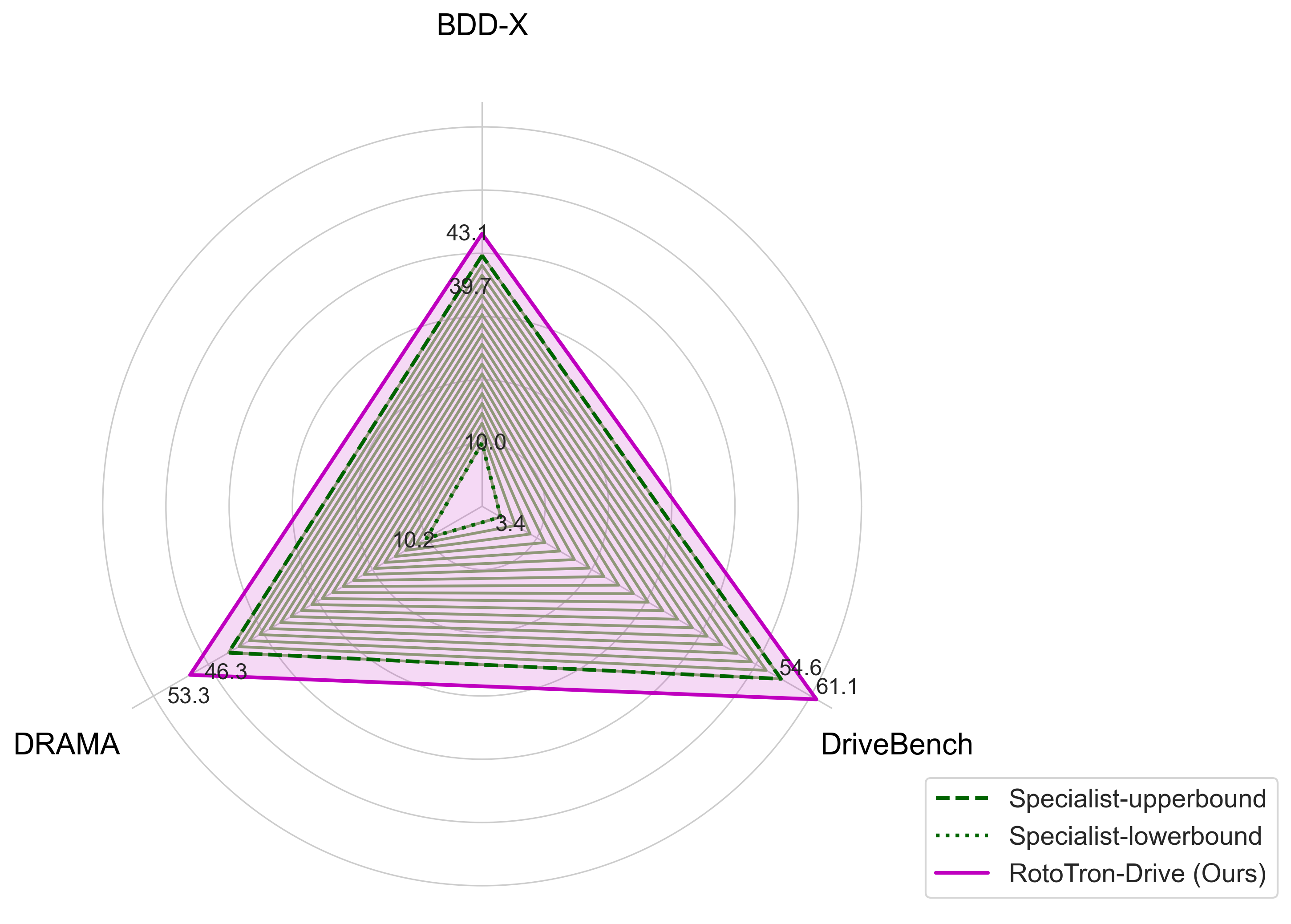}
   \end{minipage}
   \vspace{-0.4cm}
   \captionof{figure}{\textbf{\name{} achieves SOTA in both general capabilities and generalization ability.} \textbf{Left}: \name{} outperforms all specific SOTA models and other general large multimodal models across all 6 datasets comprising 13 tasks; \textbf{Right}: In zero-shot learning on unseen datasets~\cite{kim2018textual, malla2023drama, xie2025drivebench}, \name{} shows stronger generalization ability compared to models trained on individual datasets.
   }
\label{fig:radar}
\end{center}%
}]

\blfootnote{$\dagger$ Equal contribution. $\ddagger$ Corresponding author.}

\begin{abstract}

Large Multimodal Models (LMMs) have demonstrated exceptional comprehension and interpretation capabilities in Autonomous Driving (AD) by incorporating large language models.
Despite the advancements, current data-driven AD approaches tend to concentrate on a single dataset and specific tasks, neglecting their overall capabilities and ability to generalize.
To bridge these gaps, we propose \textbf{\name{}}, a general large multimodal model designed to process diverse data inputs, such as images and multi-view videos, while performing a broad spectrum of AD tasks, including perception, prediction, and planning.
Initially, the model undergoes curriculum pre-training to process varied visual signals and perform basic visual comprehension and perception tasks. 
Subsequently, we augment and standardize various AD datasets to fine-tune the model, resulting in an all-in-one LMM for autonomous driving.
To assess the general capabilities and generalization ability, we conduct evaluations on six public benchmarks and undertake zero-shot transfer on three unseen datasets, where \name{} achieves state-of-the-art performance across all tasks.
We hope \name{} as a promising solution for AD in the real world.

\end{abstract}    
{
\begin{table*}[!t]
\small
\centering
\setlength{\tabcolsep}{4.5pt} 
 \begin{tabularx}{1.0\textwidth}{l|l|cccccccc|cc|ccc} 

\toprule


\multirow{2}{*}{\centering Dataset} & \multirow{2}{*}{Type} & \multicolumn{8}{c|}{Perception} & \multicolumn{2}{c|}{Prediction} & \multicolumn{3}{c}{Planning} \\
\cmidrule(){3-15}

&  & \makecell{Scene\\Und.} & \makecell{Region\\Und.} & \makecell{Key\\Und.} & \makecell{Corner \\Und.} & \makecell{Road \\Und.} & \makecell{Risk \\Det.} & \makecell{Key \\Det.} & \makecell{Key\\ Gro.} & \makecell{Status \\Pre.} & \makecell{Motion\\ Pre.} & \makecell{Action\\Dec.}& \makecell{Driving \\Res.} & \makecell{Motion \\Pre.} \\

\midrule

CODA-LM\cite{li2024automated} & S.I.& \xmark & \cmark & \xmark & \cmark & \xmark & \xmark & \xmark & \xmark & \xmark & \xmark & \cmark & \cmark & \xmark\\
MAPLM\cite{cao2024maplm} & M.I.& \xmark & \xmark & \xmark & \xmark & \cmark & \xmark & \xmark & \xmark & \xmark & \xmark & \xmark & \xmark & \xmark\\
DriveLM\cite{sima2024drivelm} & M.I.& \cmark & \cmark & \cmark & \xmark & \xmark & \xmark & \cmark & \cmark & \cmark & \xmark & \cmark & \cmark & \xmark \\
LingoQA\cite{marcu2024lingoqa} & S.V.& \cmark & \cmark & \xmark & \xmark & \xmark & \xmark & \xmark & \xmark & \xmark & \xmark & \cmark & \cmark & \xmark\\
OmniDrive\cite{wang2024omnidrive} & M.V.& \cmark & \cmark & \cmark & \xmark & \xmark & \xmark & \xmark & \cmark & \xmark & \xmark & \cmark & \cmark & \cmark\\
NuInstruct\cite{ding2024holistic} & M.V.& \xmark & \xmark & \xmark & \xmark & \xmark & \cmark & \cmark & \cmark & \cmark & \cmark & \xmark & \cmark & \cmark\\


\bottomrule
\end{tabularx}
\vspace{4pt}
\caption{\textbf{Comparison of different AD datasets.}
Different datasets encompass various input types and multiple sub-tasks of perception, prediction, and planning in real-world scenarios.
S.I.=Single-view image, M.I.=Multi-view images, S.V.=Single-view video, M.V.=Multi-view videos.
Und.=Understanding, Det.=Detection, Gro.=Grounding, Pre.=Predition, Dec.=Decision, Res.=Resoning.}
\label{tab:data}
\vspace{-5mm}
\end{table*}
}

\section{Introduction}
\label{sec:intro}

Recently, vision-language driving datasets and models have garnered significant attention in the field of autonomous driving~\cite{chen2024survey,yang2023survey,zhang2024survey}.
Numerous datasets~\cite{sima2024drivelm, li2024automated, marcu2024lingoqa, wang2024omnidrive, kim2018textual, cao2024maplm, ding2024holistic} have been meticulously developed and curated to fine-tune Large Multimodal Models (LMMs), enabling them to better understand and generate multimodal content, as well as adapt to specific domains and applications.
Meanwhile, several methods~\cite{mao2023gptdriver, mao2023gptdriver2, pan2024vlp, wang2023bevgpt, wang2023drivemlm, wen2023dilu, li2025cross, xu2024drivegpt4, huang2025making, cui2024drive, shao2024lmdrive} have attempted to incorporate the extensive world knowledge and strong logical reasoning capabilities of Large Language Model (LLM) into AD systems, demonstrating significant improvements in interpretability and overall system performance.
Typically, these methods utilize a pre-training and fine-tuning approach\cite{liu2024visual}, where LMMs are fine-tuned on certain datasets and evaluated on the tasks within.

Due to the complexity and diversity of driving scenarios\cite{zhang2023method, ding2023surveyscenario, liu2019towards} and driver behaviors\cite{cheng2023behavexplor,mirman2019dynamical}, existing LMMs and AD datasets often focus on specific scenes and tasks\cite{li2022expansion,huang2023fuller}, as shown in Tab.~\ref{tab:data}. 
We observe that the methodologies employed in the collection of various datasets are tailored to the specific tasks they are designed to address. 
For instance, the CODA-LM dataset~\cite{li2024automated}, which is centered on corner case perception, necessitates only front view images. 
In contrast, the NuInstruct dataset which is derived from the nuScenes dataset\cite{caesar2020nuscenes}, encompasses tasks related to prediction and decision-making, necessitating multi-view or video inputs.
Furthermore, each dataset is annotated to address distinct sub-tasks; for example, MapLM\cite{cao2024maplm} is dedicated to road-related perception tasks, whereas LingoQA\cite{marcu2024lingoqa} is oriented towards planning tasks, emphasizing the action decision and driving reasoning of the ego vehicle.
As illustrated in Fig.~\ref{fig:radar}, previous specialist LMMs~\cite{li2024automated,cao2024maplm,li2024driving,marcu2024lingoqa,wang2024omnidrive,ding2024holistic} fine-tuned on these single datasets lack the general capability needed to handle the complex and varied tasks found in real-world scenarios (Left) and exhibit poor generalization performance on another dataset (Right).
This limitation highlights the need for a more general LMM to improve the versatility and robustness of AD tasks.

Therefore, we motivate an all-in-one general-purpose LMM, which is capable of simultaneously accepting various types of data inputs (\eg,~images, videos) and performing a wide range of tasks in AD (\eg,~perception, prediction, planning).
To begin with, we re-engineer an LMM to accept perspective-aware visual signals, by providing an explanation about the camera perspective and data type in the instruction. 
It allows the model to recognize the spatial relationships of objects and analyze the full context of dynamic driving environments.
To effectively train the all-in-one LMM, we employ a curriculum principle~\cite{liu2024let} to pre-train and fine-tune the model. 
This approach gradually guides the model to handle intricate data inputs, progressing from single image to multi-view videos, as well as diverse tasks, transitioning from image captioning to driving reasoning. 
In the pre-training phase, we begin by equipping LLM with a foundational ability to comprehend images through training it on image-text pairs.
To obtain a robust foundational LMM, we further conduct multi-capability pre-training, leveraging diverse types of multimodal and perception data to enhance the model's visual reasoning and perception capacities in different scenarios.
During the fine-tuning stage, we gather various open-source multimodal AD datasets presented in Tab.~\ref{tab:data}, and then enhance their mutual improvement by augmenting and standardizing their question-answer pairs.
By integrating and leveraging these diverse data and tasks, our model, \textbf{\name{}}, is capable of efficiently performing various AD tasks in real-world scenarios.
Experimentally, we thoroughly evaluate out \name{} on all challenging benchmarks, where it achieves state-of-the-art performance across all tasks.

To summarize, our contributions are:
\begin{itemize}
    \item We propose a novel all-in-one large multimodal model, \name{}, robustly equipped with the general capabilities to execute a range of AD tasks and the generalization ability to effectively transfer to new datasets.
    \item We introduce comprehensive benchmarks for evaluating autonomous driving LMMs, which include six public datasets, four input types, and thirteen challenging tasks. 
    To the best of our knowledge, this is the first to use multiple benchmarks to evaluate autonomous driving LLMs. 
    \item We present a curriculum principle for pre-training and fine-tuning on both diverse multimodal data and AD data.
    \name{} demonstrates state-of-the-art performances and consistently outperforms models trained on the individual dataset across all evaluated benchmarks. 
\end{itemize}

\begin{figure*}[ht]
  \includegraphics[width=\linewidth]{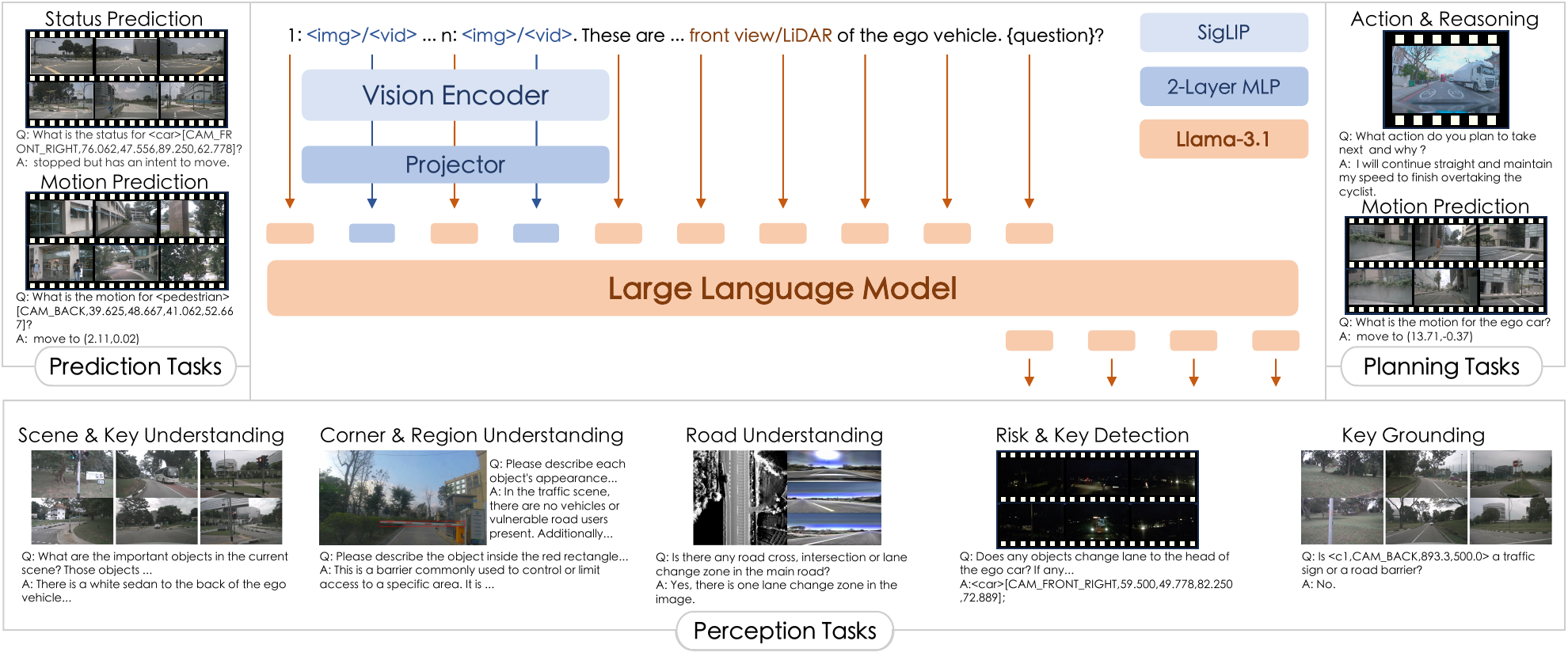}
\vspace{-6mm}
  \caption{\textbf{Overview of \name{} framwork.}
We adapt the architecture form of LLaVA~\cite{liu2024visual} with a different model instantiation, processing various visual input signals.
We design a perspective-aware prompt to accept multi-perspective inputs in AD scenario.
Equipped with diverse AD multimodal data, \name{} possesses an all-in-one capability to accomplish multiple tasks in autonomous driving.}
\vspace{-2mm}
\label{fig:pipeline}
\end{figure*}

\vspace{-2mm}
\section{Related Work}
\label{sec:related}
\subsection{Vision-Language Driving Datasets}
In recent years, numerous vision-language driving datasets have been developed with the aim of training and evaluating LMMs designed for AD scenarios.
~\cite{kim2018textual,xu2020explainable,xu2024drivegpt4,park2024vlaad} are dedicated to scene description and actions decision in driving videos. DRAMA~\cite{malla2023drama}, CODA-LM~\cite{li2024automated}, and DriveVLM~\cite{tian2024drivevlm} focus on risk objects and corner cases learning.
In addition to the single-view data, many studies construct multi-view data based on the nuScenes dataset~\cite{caesar2020nuscenes}. 
For instance, NuScenes-QA~\cite{qian2024nuscenes} introduces free-form question-answer annotations for 3D object relationships. 
DriveLM~\cite{sima2024drivelm}, OmniDrive~\cite{wang2024omnidrive},
and NuInstruct~\cite{ding2024holistic} employ the original annotations and LLMs to generate visual question-answer pairs covering perception, reasoning, and planning. 
Furthermore, MAPLM~\cite{cao2024maplm} integrates multi-view data and LiDAR data to analyze and recognize road surface conditions. 
In this paper, we augment and standardize multiple driving datasets to train a comprehensive LMM for diverse autonomous driving scenarios.

\subsection{LMMs for Autonomous Driving}
LMMs have demonstrated impressive performance in diverse tasks~\cite{liu2024visual,achiam2023gpt,luo2024autom3l,liu2024robouniview,chen2024timemarker,hwang2024emma,yan2024robomm,zeng2025distime,zhong2025p3nav}.
Recently, researchers have begun to explore the potential of LLMs in the field of AD. 
DiLu~\cite{wen2023dilu} and GPT-Driver~\cite{mao2023gpt} attempt to utilize GPT-3.5 and GPT-4 as driving planners. 
Subsequently, DriveGPT4~\cite{xu2024drivegpt4} and RDA-Driver~\cite{huang2025making} introduce end-to-end LMMs that generate control signals or trajectories. 
Unlike the methods that handle driving maneuvers through language, LMDrive~\cite{shao2024lmdrive} and DriveMLM~\cite{wang2023drivemlm} use a decoder to predict control signal from hidden embeddings.
In order to enhance the perception and reasoning abilities, several approaches aim to improve the model architecture.
Reason2Drive~\cite{nie2025reason2drive} proposes a prior tokenizer to extract local image features and BEV-InMLLM~\cite{ding2024holistic} injects Bird's-Eye-View (BEV) representations into LMMs. 
OmniDrive~\cite{wang2024omnidrive} uses Q-Former3D to integrate 2D pre-trained knowledge with essential 3D spatial understanding.
ELM~\cite{zhou2024embodied} incorporates a time-aware token selection module to accurately inquire about temporal cues. 
Although these methods have demonstrated satisfactory performance, their applicability is limited to the specific scene and task, such as a particular data type or a dataset-specific task. 
In light of this, we propose an all-in-one LMM designed to efficiently process diverse driving scenes and tasks in AD.

\section{Methodology}
\label{sec:methodology}

\subsection{Overview}
\label{sec-overview}
We propose \name, illustrated in Fig.~\ref{fig:pipeline}, an all-in-one LMM designed to efficiently process various driving data and tasks.
Given a visual signal $X_{\text{v}}$ captured by the sensors and a user instruction $X_{\text{t}}$, \name~$\mathcal{F}(\cdot)$ provides the driving-related analysis and suggestions:
\begin{equation}
Y_{\text{t}} = \mathcal{F}(X_{\text{v}}, X_{\text{t}}).
\end{equation}
$X_{\text{v}}$ can represent various data formats, including image, multi-images, video, and multi-videos captured by a single camera, multi-view cameras, or LiDAR, while $X_{\text{t}}$ encompasses questions pertaining to perception, 
prediction, and more.
By integrating diverse data and tasks, \emph{\name~can be trained on a wide range of AD vision-language data, resulting in mutual improvements across different datasets and tasks}.
Moreover, \emph{once trained, \name~can be effectively deployed across a broad spectrum of real-world AD scenarios}, \eg~different camera and radar system configurations, as well as various AD tasks.

In the following sections, we first describe the architecture of \name, which has the capability to process multiple types of data captured by different sensors (Sec.~\ref{sec-model}).
To facilitate the model's comprehension of AD scenarios, we gather diverse datasets with multiple data formats and tasks, then augment and standardize their question-answer pairs to enhance collaboration across different datasets (Sec.~\ref{sec-data}).
In order to effectively train \name, we adopt a curriculum learning approach to progressively enhance the model's capability (Sec.~\ref{sec-training}).

\subsection{Model Architecture}
\label{sec-model}
\paragraph{Preliminary.}
The predominant LMMs comprise three components: a vision encoder, a projector and a LLM.
In particular, the vision encoder encodes the input images into the visual features.
Afterward, the projector projects the image features into the word embedding space.
Based on the visual tokens and the user instruction, the LLM computes the probability of the target word step by step.

\vspace{-4mm}
\paragraph{Perspective-aware prompt.} 
Typical LMMs~\cite{liu2024visual,bai2023qwen} flatten visual features for LLM input, failing to distinguish between perspectives (e.g., front or back view) and formats (e.g., image or video).
To address this, we propose a perspective-aware prompt.
As shown in the Tab.~\ref{tab:prompt}, we utilize different placeholders for image and video inputs, where the placeholders will be replaced by respective tokens before being fed into LLM. 
We also assign numerical labels with different perspectives and explain the specific camera or LiDAR for each in the text.
In order to enhance computational efficiency, we apply a $2\times2$ spatial pooling on the video features, and flatten them into the visual tokens.
{
\begin{table}[t]


\begin{tcolorbox}[colback=gray!10,
                  colframe=black,
                  width=\linewidth,
                  arc=1mm, auto outer arc,
                  boxrule=0.5pt,
                 ]
\small
\vspace{-1mm}
\textbf{\textbf{Perspective-aware Prompt:}} 

\vspace{1mm}
1: \textcolor{Blue}{\textless image\textgreater/\textless video\textgreater}\;2: \textcolor{Blue}{\textless image\textgreater/\textless video\textgreater} ... n: \textcolor{Blue}{\textless image\textgreater/\textless video\textgreater}. These n \textcolor{Blue}{images/videos} are the \textcolor{BurntOrange}{front view}, \textcolor{BurntOrange}{front left view}, ..., and \textcolor{BurntOrange}{LiDAR} of the ego vehicle. \{question\}?
\vspace{-1mm}
\end{tcolorbox}
\vspace{-2mm}
\caption{The perspective-aware prompt for multi-view inputs.
}
\vspace{-6mm}
\label{tab:prompt}
\end{table}
}

\begin{figure*}[ht]
  \includegraphics[width=\linewidth]{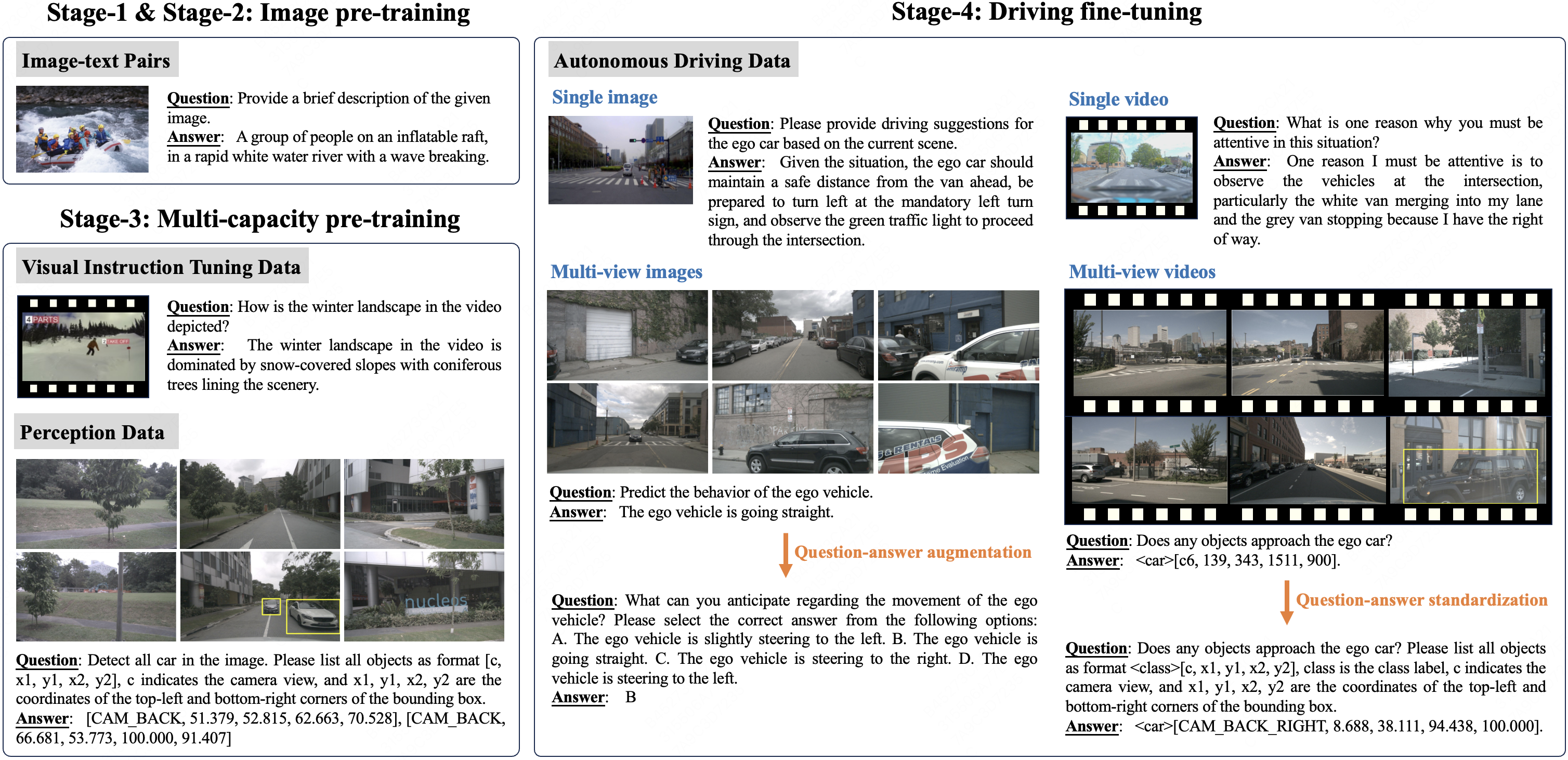}
  \vspace{-4mm}
  \caption{\textbf{Illustration of the curriculum learning framework.} \textbf{Stage-1 \& Stage-2}: it consists of language-image alignment and single-image pre-training, which use the image-text pairs to equip LLM with a foundational capability for single-image comprehension. We refer to the combination of these two stages as image pre-training. \textbf{Stage-3}: we enhance the model’s visual reasoning and perception capabilities across diverse scenarios by training on both the visual instruction tuning data and perception data. \textbf{Stage-4}: we further fine-tune the model on six augmented and standardized autonomous driving datasets, enabling \name~to tackle a 
  wide range of AD tasks.
  }
\label{fig:data_example}
\vspace{-5mm}
\end{figure*}

\subsection{Data}
\label{sec-data}
In the training of LMMs, data plays a crucial role in enabling and activating the LLM's ability to understand multimodal information. 
To enhance the comprehension and reasoning skills of \name~in multimodal AD scenarios, we construct three datasets: conventional multimodal data, perception data, and autonomous driving data.

\subsubsection{Conventional Multimodal Data}
\label{sec:m-data}
Recent studies~\cite{chen2024internvl,bai2023qwen} show that LMMs can achieve enhanced performance as the volume of data increases. 
However, compared to the abundant image-text data available online~\cite{sharma2018conceptual,schuhmann2022laion}, AD image-text data is significantly limited. 
To enhance the performance of \name, we pre-train a base model with extensive multimodal data, enabling reasoning with single images, multi-images, and video.

Specifically, we construct a multimodal dataset from~\cite{li2024llava} comprising image-text pairs and diverse visual instruction tuning data.
The objective of the image-text pairs is to align the vision encoder and LLM, enabling the model to develop a foundational understanding of images. 
We utilize multiple datasets, including LCS-558K~\cite{liu2024visual}, COCO118K~\cite{li2024llava}, CC3M~\cite{sharma2018conceptual}. 
To enhance the model's capability in dealing with the visual data in various sensor configurations such as single-view and multi-view cameras, we utilize visual instruction tuning data in the OneVision data~\cite{li2024llava}, including image, multi-images and video.

\subsubsection{Perception Data}
\label{sec:p-data}
To equip \name~with AD perception capabilities, we create a comprehensive grounding dataset including various data formats.
For the single image data, we utilize the COCO~\cite{lin2014microsoft} and Object365~\cite{shao2019objects365} datasets. 
We randomly select a category from an image and use grounding prompts (e.g., ``Detect all \texttt{<category>} in the image.") to prompt the model to detect all objects in that category.
We represent the object's position with the bounding box $[x_{min}, y_{min}, x_{max}, y_{max}]$ or the region center $[x_{center}, y_{center}]$. 
The $x$ and $y$ values are normalized in a range of 0 to 100 based on the image's size.
For the multi-view images and multi-view videos, we employ the nuScenes~\cite{caesar2020nuscenes} dataset. 
To imbue the model with a sense of spatial awareness, we expect it not only to predict the object bounding boxes but also to estimate the camera perspective. 
Therefore, we represent the object's position with $[cam, x_{min}, y_{min}, x_{max}, y_{max}]$ or $[cam, x_{center}, y_{center}]$, where $cam$ denotes the camera perspective such as ``CAM\_BACK".
An example of the perception data is illustrated in the bottom left of Fig.~\ref{fig:data_example}.

\subsubsection{Autonomous Driving Data}
\label{sec:d-data}
Here we collect diverse datasets to train an all-in-one LMM that can synchronously tackle various AD tasks in different scenarios. 
Concretely, we use six autonomous driving datasets: CODA-LM~\cite{li2024automated}, MAPLM~\cite{cao2024maplm}, DriveLM~\cite{sima2024drivelm}, LingoQA~\cite{marcu2024lingoqa}, OmniDrive~\cite{wang2024omnidrive} and NuInstruct~\cite{ding2024holistic}. The detailed descriptions of the six datasets are shown in Tab.~\ref{tab:data}. 
These datasets encompass various sensor configurations such as camera and LiDAR, and different AD tasks including perception, prediction, and planning. 
It is noteworthy to mention that different datasets may exhibit distinct question patterns. To foster collaborative enhancement, we augment and standardize the question-answer pairs as follows.

\vspace{-5mm}
\paragraph{Question-answer augmentation.} Some datasets are restricted to a fixed set of templates. 
For instance, CODA-LM comprises only three question templates, while MAPLM utilizes five. 
It hinders the potential for the model's generalization. 
To overcome this limitation, we employ GPT-4o-mini to augment the question-answer pairs and increase their diversity.
Additionally, a significant portion of the questions are open-ended. 
To further enhance the diversity, we randomly transform some open-ended questions into multiple-choice format. 
An example of the augmentation is illustrated in the bottom right of Fig.~\ref{fig:data_example}.
Please refer to the supplementary material for
more details.
\vspace{-4mm}
\paragraph{Question-answer standardization.} 
Different datasets may exhibit inconsistencies in question-answer styles. 
For example, DriveLM uses ``\texttt{<}c6, CAM\_BACK, 1088.3, 497.5\texttt{>}" to represent an object, where ``c6" denotes the class ID.
In contrast, NuInstruct employs the format of ``\texttt{<}car\texttt{>}[c6, 139, 343, 1511, 900]", where ``c6" represents the camera ID. 
To ensure compatibility across datasets, we standardize the representation of objects and explicitly specify the representation format.
Moreover, to accommodate bounding boxes in images of different sizes, we standardize the coordinates of the bounding boxes to a range of 0 to 100 based on the image's size.
For example, for the NuInstruct dataset, we re-represent the object as ``\texttt{<}car\texttt{>}[CAM\_BACK\_RIGHT, 8.688, 38.111, 94.438, 100.000]" 
and add the formatting instructions at the end of the question, as illustrated in the bottom right of Fig.~\ref{fig:data_example}.

{
\begin{table*}[!t]
\small
\setlength{\tabcolsep}{2.4pt} 
\centering
 \begin{tabularx}{1.0\textwidth}{l|l|ccc|ccc} 
\toprule

\multirow{2}{*}{Dataset} & \multirow{2}{*}{Metrics} &
\multicolumn{3}{c|}{Zero-shot Models} & \multicolumn{3}{c}{Fine-tuned Models}
\\
\cmidrule{3-8}
& & GPT-4o~\cite{hurst2024gpt4o} &  LLaVA-OV~\cite{li2024llava} & \name{}-Zero & Specialist Model$^{\dag}$ & Drive-OV & \textbf{\name{}}\\
\midrule

\multirow{4}{*}{CODA-LM~\cite{li2024automated}} 
& General$\uparrow$  & 47.06 & 38.70 & 34.74 & \textbf{55.04} & 53.60 &52.94\\
& Regional$\uparrow$ & 50.37& 51.70 & 54.46 & 77.68 & 77.64&\textbf{77.76}\\
& Suggestion$\uparrow$ & 48.94 & 49.32 & 38.92 & 58.14 & 58.72&\textbf{61.84}\\
\rowcolor{mygray} & Average $\uparrow$ & 48.79 & 46.57 & 41.70 & 63.62 & 63.32 &\textbf{64.18}\\
\midrule

\multirow{3}{*} {MAPLM~\cite{cao2024maplm}} 
& FRM$\uparrow$ & 0.33 &0.00 & 0.00 &57.99 & 62.33& \textbf{64.80}\\
& QNS$\uparrow$ & 45.9 & 36.18& 42.27 & 85.52 & 87.55& \textbf{88.53}\\
\rowcolor{mygray} & Average$\uparrow$ & 23.12& 18.09& 21.13 & 71.76 & 74.94 & \textbf{76.67}\\
\midrule

\multirow{5}{*} {DriveLM~\cite{sima2024drivelm}}  
& Accuracy$\uparrow$ & 38.55 & 25.03 & 52.94 & 73.39 & \textbf{79.38} & 76.09\\
& ChatGPT$\uparrow$  & 67.27 & 65.70 & 51.11 & 65.25 & 65.92& \textbf{66.44}\\
& Language$\uparrow$ & 8.97  & 14.44 & 7.12 & 48.56 & 47.28 & \textbf{48.90}\\
& Match$\uparrow$    & 24.00 & 40.93 & 35.05 & 47.65 & 40.70& \textbf{48.63}\\
\rowcolor{mygray} & Average $\uparrow$ & 41.21 & 42.36 & 39.47 & 60.02 & 59.84 & \textbf{61.30}\\
\midrule

\rowcolor{mygray} LingoQA~\cite{marcu2024lingoqa} 
 & Lingo-Judge$\uparrow$ & 18.40 & 38.80& 38.60 & 60.80 & \textbf{70.10} &69.20\\
\midrule

\multirow{4}{*} {OmniDrive~\cite{wang2024omnidrive}}
& BLEU$\uparrow$  & 10.91  & 16.14 & 20.30 & 38.00 & 38.25 & \textbf{39.11}\\
& CIDEr$\uparrow$  & 24.42 & 28.41 & 34.33 & 68.60 & 76.04 & \textbf{77.50} \\
& ROUGE$\uparrow$  & 22.34 & 22.14 & 23.67 & 32.60 & 33.36 & \textbf{34.15} \\
\rowcolor{mygray} & Average $\uparrow$ & 19.22 & 22.23 & 26.10 & 46.40 & 49.22 & \textbf{50.25}\\
\midrule

\multirow{5}{*} {NuInstruct~\cite{ding2024holistic}}
& MAE$\downarrow$    & 9.93 & 87.04 & 19.36 & 9.08 & 1.81 & \textbf{1.56}\\
& Accuracy$\uparrow$ & 10.64& 3.75 & 2.57 & 32.48 & 64.57 &\textbf{64.71} \\
& MAP$\uparrow$      & 0.00 & 0.00 & 0.00 & 21.93 & 23.39 &\textbf{39.04} \\
& BLEU$\uparrow$     & 7.08 & 8.55 & 8.06 & 35.20 & 80.85 &\textbf{83.00}\\
\rowcolor{mygray} & Average$^{*}$ $\uparrow$ & 1.95 & 0.00 & 0.00 & 20.13 & 41.75 &\textbf{46.30}\\



\bottomrule
\end{tabularx}

\caption{General performance on benchmarks.
We compare with state-of-the-art specialist models, commercial models and open-source large multimodal models across diverse autonomous driving valuation benchmarks spanning multiple modalities.
$^{\dag}$Specialist models correspond to the performance of six different
models~\cite{li2024automated,cao2024maplm,li2024driving,marcu2024lingoqa,wang2024omnidrive,ding2024holistic}.
$^{*}$ indicates \texttt{max((Accuracy+MAP+BLEU-MAE)/4, 0)}.
}
\label{tab:sota}
\end{table*}
}

\subsection{Training}
\label{sec-training}
In this section, we present a curriculum learning approach to progressively improve the performance of the model on various AD data and tasks, resulting in an all-in-one autonomous driving model \name. Specifically, we gradually increase the complexity of the data, progressing from a single image to multiple videos, and the complexity of the tasks, transitioning from image captioning to driving reasoning, for training \name. As illustrated in Fig.~\ref{fig:data_example}, the training process is divided into four steps:
\vspace{-4mm}
\paragraph{Stage-1: Language-image alignment.} The goal of this stage is to equip the pre-trained LLM with a foundational capability for multimodal comprehension. To achieve that, we train the projector to align with the word embedding space of LLM. 
We froze both the vision encoder and LLM, and only optimize the projector on LCS-558K~\cite{liu2024visual}.
\vspace{-4mm}
\paragraph{Stage-2: Single-image pre-training.} In this stage, we further enhance the model's capacity to comprehend single-image by collectively optimizing the entire model. 
We use the image-text pairs outlined in Sec.~\ref{sec:m-data}, and optimize all the parameters of the model to enhance the suitability of LLM for multimodal tasks.
\vspace{-4mm}
\paragraph{Stage-3: Multi-capacity pre-training.} To obtain a robust foundational model for training AD systems, we enhance the model's reasoning and perception capabilities across diverse scenarios. 
To this end, we utilize the visual instruction tuning data described in Sec.~\ref{sec:m-data} to enhance the model to reason about fundamental visual elements. 
Additionally, we employ the perception data described in Sec.~\ref{sec:p-data} to facilitate the model's perception capacity.
It is noteworthy that the training data comprises diverse data formats, including single image, single video, multi-view images, and multi-view videos. By equipping the model with the capability to process various data and tasks, we establish the groundwork for training an all-in-one AD model. 
\vspace{-4mm}
\paragraph{Stage-4: Driving fine-tuning.} To enable \name~to tackle a wide range of AD tasks, we further fine-tune the model on diverse driving datasets. Specifically, we utilize six augmented and standardized autonomous driving datasets outlined in Sec. \ref{sec:d-data}. In this stage, we optimize all the parameters of the model. Once trained, the proposed all-in-one \name~can be effectively deployed across a broad spectrum of AD scenarios, \eg~different camera and radar system configurations, as well as various AD tasks.

\vspace{-2mm}
\section{Experiment}

{
\begin{table*}[!t]
\small
\setlength{\tabcolsep}{6.3pt} 
\centering
 \begin{tabularx}{1.0\textwidth}{l|l|cccccc|c} 
\toprule
\multirow{2}{*}{Dataset} & \multirow{2}{*}{Metrics} &
\multicolumn{6}{c|}{Specialist Model w/} & Generalist Model
\\
\cmidrule(){3-9}
& & \makecell{CODA-LM} & \makecell{MAPLM} & \makecell{DriveLM} & \makecell{LingoQA} & \makecell{OmniDrive} & \makecell{NuInstruct} & \name{} \\
 
\midrule

BDD-X\cite{kim2018textual} & \multirow{3}{*}{GPT-Score} & 36.40 & 10.01 & 35.04 & 28.93 & 39.67 & 26.48  & \textbf{43.10} \\
DRAMA\cite{malla2023drama} & & 46.30 & 10.15 & 28.92 & 45.68 & 46.20 & 10.52 & \textbf{53.32}\\
DriveBench\cite{xie2025drivebench} & & 31.82 & 12.92 & 54.48 & 28.63 & 39.14 & 3.39 & \textbf{61.06}\\

\bottomrule
\end{tabularx}
\caption{Generalization ability on unseen datasets. 
Specialists are fine-tuned on a single 
dataset, whereas \name{} is on all.
}
\label{tab:zero-shot}
\end{table*}
}
{
\begin{table*}[!t]
\vspace{-2mm}
\small
\setlength{\tabcolsep}{8.2pt} 
\centering
\begin{tabularx}{1.0\textwidth}{l|cccccc|c|c} 
\toprule


Task & \makecell{Dolphins\\-7B~\cite{ma2024dolphins}} & \makecell{DriveLM\\-7B~\cite{sima2024drivelm}} & \makecell{Qwen2-VL\\-7B~\cite{wang2024qwen2vl}} & \makecell{InterVL2\\-8B~\cite{chen2024internvl}} & \makecell{Qwen2-VL\\-72B~\cite{wang2024qwen2vl}} & GPT-4o~\cite{achiam2023gpt} & \textcolor{black!50}{Human} & \makecell{\name{}\\-8B} \\
\midrule
Perception & 9.59 & 16.85 & 28.99 & 32.36 & 30.13 & 35.37 & \textcolor{black!50}{47.67} & \textbf{41.11} \\
Prediction & 32.66 & 44.33 & 37.89 & 45.52 & 49.35 & 51.30 & \textcolor{black!50}{-} & \textbf{57.95} \\
Planning & 52.91 & 68.71 & 57.04 & 53.27 & 61.30 & 75.75 & \textcolor{black!50}{-} & \textbf{79.44} \\
Behavior & 8.81 & 42.78 & 49.07 & 54.58 & 51.26 & 45.40 & \textcolor{black!50}{69.51} & \textbf{62.85} \\ 
\rowcolor{mygray} Average $\uparrow$ & 25.99 & 43.17 & 43.25 & 46.43 & 48.01 & 51.96 & \textcolor{black!50}{-} & \textbf{61.06}\\

\bottomrule
\end{tabularx}
\vspace{-4pt}
\caption{Performance comparison of LMMs 
across different tasks on 
DriveBench~\cite{xie2025drivebench}. \name{} outperforms across all tasks.}
\label{tab:drivebecnch}
\vspace{-4mm}
\end{table*}
}

\subsection{Experimental Setting}
\paragraph{Dataset and Evaluation Metrics.}
We utilize six open-source autonomous driving datasets: CODA-LM~\cite{li2024automated}, MAPLM~\cite{cao2024maplm}, DriveLM~\cite{sima2024drivelm}, LingoQA~\cite{marcu2024lingoqa}, OmniDrive~\cite{wang2024omnidrive} and NuInstruct~\cite{ding2024holistic}.
DriveLM, OmniDrive and Nuinstruct are annotated based on the nuScenes\cite{caesar2020nuscenes} dataset, suitable for multi-view images or videos input.
The CODA-LM dataset, designed for corner case question-answer pairs, serves as a single-view dataset.
MAPLM is a multi-view image dataset and LingoQA is a a single-view video dataset.
We follow the common practice metrics in each work for fair comparison:
(1) CODA-LM uses text-only LLMs, \eg~GPT-4, as evaluators to score model responses by few-shot learning.
(2) MAPLM evaluates fine-grained QAs with multi-class classification correct ratio as the accuracy metric, while open QAs are assessed with a rule-based BLEU metric.
(3) DriveLM implements four evaluation methods: accuracy, LLM score, language rule-based evaluation, and match score.
(4) LingoQA uses a learned text classifier Lingo-Judge to estimate the score of model answers.
(5) OmniDrive employs rule-based language metrics to evaluate sentence similarity at the word level. 
(6) NuInstruct uses a variety of metrics to evaluate different tasks: Mean Absolute Error (MAE) for regression tasks, accuracy for classification tasks, Mean Average Precision (MAP) for detection tasks, and a rule-based BLEU metric for captioning tasks.

\vspace{-6mm}
\paragraph{Training configurations.}
We adapt SigLIP~\cite{zhai2023sigmoid} as the vision encoder with a
resolution of 384$\times$384 and a 2-layer MLP~\cite{liu2024improved} as the projector to project the image features into the word embedding space.
For the language model, we choose Llama-3.1 8B~\cite{meta2024introducing}, which uses a tokenizer with a vocabulary of 128K tokens. 
Our model is trained on sequences of 8,192 tokens.
We conduct our experiments using 32 A100 GPUs. 
The four stages require 1h, 26h, 90h, and 60h, respectively.
More details regarding experimental setup can be found in supplementary materials.

\subsection{Main Results}
\begin{figure}[t]
    \centering
    \vspace{-1mm}
    \includegraphics[width=1.0\linewidth]{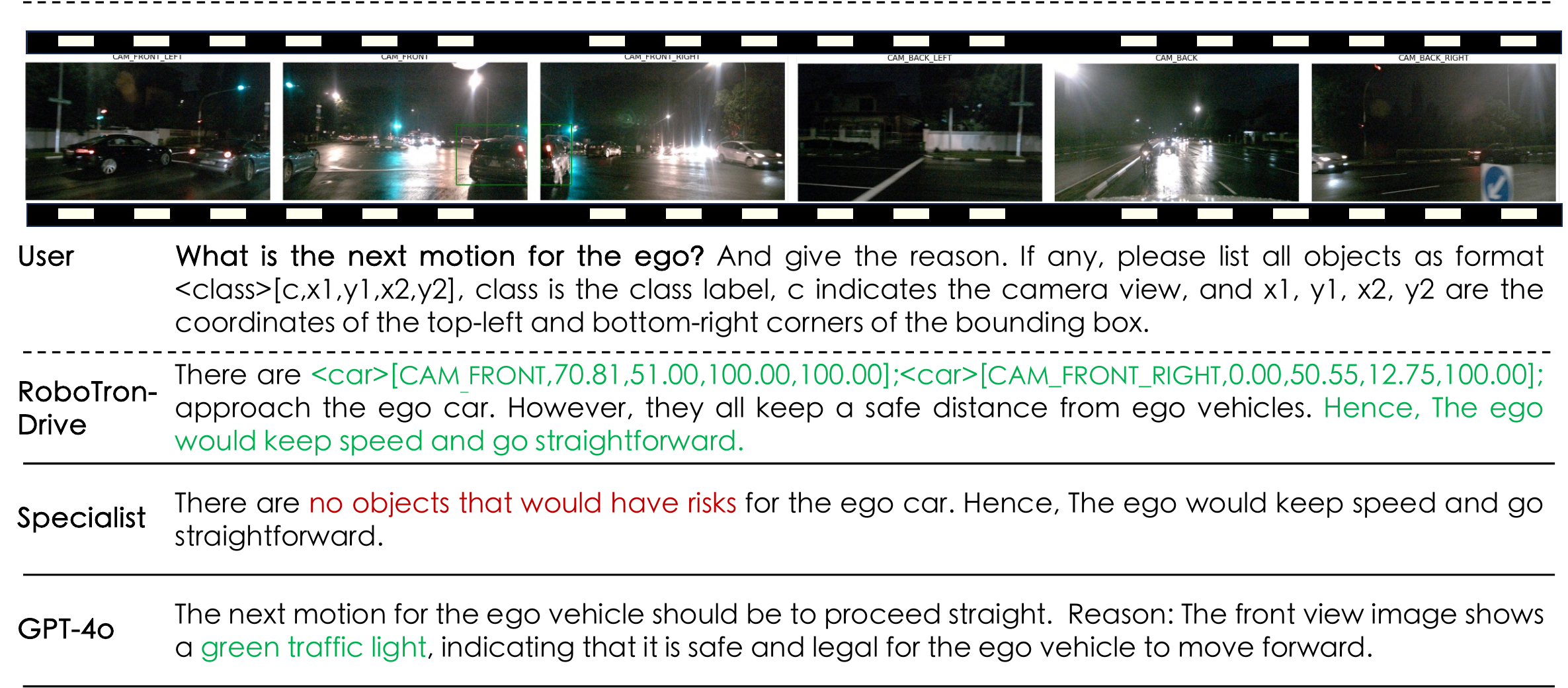} 
    \vspace{-7mm}
    \caption{An example of how our \name{} model responds to user queries based on multi-view video inputs. Compared to specialist and public models, our model successfully understands the surrounding environment and makes accurate decisions, ultimately outputting the textual answer in the user-specified format.} 
    \label{fig:fig_example} 
    \vspace{-5mm}
\end{figure}

\vspace{-2mm}
\paragraph{General capabilities.}
We compare \name{} with current state-of-the-art models and report the results in all six benchmarks. 
As depicted in Tab.~\ref{tab:sota}, \name{} surpasses the previous works across all benchmarks, achieving remarkable improvements such as \textbf{+4.91} in MAPLM and \textbf{+26.17} in NuInstruct.
It is worth noting that specialist LMMs trained on specific datasets have shown excellent performance on specific tasks.
However, by utilizing extensive multi-type information from multiple datasets, \name{} not only simultaneously accomplishes all tasks in real scenarios but also outperforms all specialist models.
This underscores our model's capability to provide a comprehensive solution by efficiently utilizing diverse AD data and tasks.
Furthermore, \name{} demonstrates superior performance compared to zero-shot models such as GPT-4o~\cite{hurst2024gpt4o}, LLaVA-OV~\cite{li2024llava}, and \name{}-Zero. 
\name{}-Zero, a variant of our model without fine-tuning on AD data, serves as a baseline. 
Since these models are not fine-tuned on multi-view videos and AD data, their performance is notably poor in related tasks, such as NuInstruct.
To validate the effectiveness of our perspective-aware prompt and AD dataset, we enhanced LLaVA-OV by equipping it with the prompt and fine-tuning it on our augmented datasets. 
The resulting model Drive-OV shows substantial improvements across all benchmarks, highlighting the compatibility of our prompt and dataset. 
Drive-OV is pre-trained on a significant amount of videos, while LingoQA is confined to simple scene understanding and action description tasks.
Consequently, Drive-OV demonstrates performance comparable to \name{} in LingoQA.
However, on more complex datasets involving detection and grounding tasks like DriveLM and NuInstruct, \name{} outperforms Drive-OV.
This highlights our multi-capability pre-trained model has greater potential to process complex AD scenes, with Fig.~\ref{fig:fig_example} illustrating a practical application in Nuinstruct.

\begin{table*}[!t]
\begin{minipage}[!t]{0.48\textwidth}
\centering
\scriptsize
\addtolength{\tabcolsep}{-5pt} 
\begin{tabular}{cc|cccccc|c}
\toprule
\makecell{QA\\aug.} & \makecell{QA\\stand.} & CODA-LM & MAPLM & DriveLM & LingoQA & OmniDrive & NuInstruct & Avg.\\
\midrule

\xmarkg & \xmarkg & 62.86 & 70.18 & 47.41 & 68.60 & 49.69 & 34.51 & 55.54 \\
\cmarkg & \xmarkg & 63.93 & 72.71 & 59.67 & 70.80 & 49.70 & 34.77 & 58.60 \\
\cmarkg & \cmarkg & 63.90 & 74.05 & 60.54 & 71.20 & 50.23 & 42.44 & \textbf{60.39} \\

\bottomrule

\end{tabular}
\vspace{-2mm}
\caption{Ablation on the proposed Question-Answer enhancement. aug.=augmentation, stand.=standardization.}
\label{tab:ablation_data}
\end{minipage}
\hfill
\begin{minipage}[!t]{0.48\textwidth}
\centering
\scriptsize
\setlength\tabcolsep{1pt} 
\begin{tabular}{cc|cccccc|c}
\toprule
CL & Pre. & CODA-LM & MAPLM & DriveLM & LingoQA & OmniDrive & NuInstruct & Avg.\\
\midrule

\xmarkg  & Mixed & 60.80 & 70.71 & 52.10 & 58.78 & 50.32 &  41.53 & 55.71\\ 
\midrule
\cmarkg & S0 & 63.91 & 74.60 & 59.35 & 64.20 & 49.68 & 40.05 & 58.63 \\
\cmarkg & S1 & 64.09 & 74.20 & 58.96 & 67.60 & 48.87 & 39.45 & 58.86 \\
\cmarkg & S2 & 63.90 & 74.05 & 60.54 & 71.20 & 50.23 & 42.44 & 60.23 \\
\cmarkg & S3 & 64.18 & 76.67 & 61.30 & 69.20 & 50.25 & 46.30 & \textbf{61.32} \\ 

\bottomrule
\end{tabular}
\vspace{-2mm}
\caption{Ablation on curriculum learning. Pre.=Pretrained.}
\label{tab:ablation_training}
\end{minipage}
\vspace{-2mm}
\end{table*}

\begin{table*}[!t]
\begin{minipage}[t]{0.48\textwidth}
\centering
\scriptsize
\setlength\tabcolsep{1pt} 
\begin{tabular}{l|cccccc|c}
\toprule
Data & CODA-LM & MAPLM & DriveLM & LingoQA & OmniDrive & NuInstruct & Avg.\\
\midrule

Individual & 64.13 & 74.02 & 60.91 & 67.40 & 49.22 & 45.06 & 60.12\\
Mixed & 64.18 & 76.67 & 61.30 & 69.20 & 50.25 & 46.30 & \textbf{61.32} \\

\bottomrule
\end{tabular}
\vspace{-2mm}
\caption{Ablation on multi-dataset fine-tuning.}
\label{tab:ablation_multidata}
\end{minipage}
\hfill
\begin{minipage}[t]{0.48\textwidth}
\centering
\scriptsize
\addtolength{\tabcolsep}{-3.5pt} 
\begin{tabular}{c|cccccc|c}
\toprule
PP & CODA-LM & MAPLM & DriveLM & LingoQA & OmniDrive & NuInstruct & Avg.\\
\midrule

\xmarkg & 63.50 & 73.87 & 61.04 & 70.20 & 49.32 & 46.18 & 60.69 \\
\cmarkg  & 64.18 & 76.67 & 61.30 & 69.20 & 50.25 & 46.30 & \textbf{61.32} \\

\bottomrule

\end{tabular}
\vspace{-2mm}
\caption{Ablation on perspective-aware prompt.}
\label{tab:ablation_prompt}
\end{minipage}
\vspace{-2mm}
\end{table*}
\begin{table}[h]
        \vspace{-7pt}
       \centering \scriptsize
       \setlength\tabcolsep{5.3pt}
       \begin{tabularx}
       {0.48\textwidth}{l|cccc|cccc}
       \toprule
       \multirow{2}{*}{Model} &
        \multicolumn{4}{c|}{L2(m)$\downarrow$} & \multicolumn{4}{c}{Collision(\%)$\downarrow$} 
        \\
        \cmidrule(r){2-5} \cmidrule(lr){6-9} 

        & 1s & 2s & 3s & Avg. & 1s & 2s & 3s & Avg.  \\
        \midrule

        UniAD~\cite{hu2023planning} & 0.20 & 0.42 & 0.75 & 0.46 & 0.02 & 0.25 & 0.84 & 0.37 \\
        VAD-Base~\cite{jiang2023vad} & 0.17 & 0.34 & 0.60 & 0.37 & 0.04 & 0.27 & 0.67 & 0.33 \\
        \name{} & 0.14 & 0.30 & 0.57 & \textbf{0.33} & 0.03 & 0.12 & 0.63 & \textbf{0.26} \\

       
       \bottomrule
       \end{tabularx}
    \vspace{-2mm}
   \caption{Comparison of planning task on nuScenes $val$ set.}
   \label{tab:ablation_planning}
   \vspace{-2mm}
\end{table}


\vspace{-5mm}
\paragraph{Generalization ability.}
To verify the model's generalization capability, we conduct zero-shot experiments on three widely-used datasets: BDD-X~\cite{kim2018textual}, DRAMA~\cite{malla2023drama}, and DriveBench~\cite{xie2025drivebench}. 
These datasets encompass perception~\cite{malla2023drama}, prediction~\cite{kim2018textual}, and planning~\cite{xie2025drivebench} tasks, providing a comprehensive evaluation framework.
Following prior works ~\cite{li2024automated, tian2024drivevlm}, we employ GPT-4o~\cite{hurst2024gpt4o} as the evaluator to assess the alignment between the model's predictions and the ground truths in BDD-X and DRAMA. 
For DriveBench, we utilize the standard evaluation metric proposed in ~\cite{xie2025drivebench}.
For comparison, we train the specialist models on each single dataset and evaluate their performance on unseen datasets by predicting task-specific outputs.
As shown in Tab.~\ref{tab:zero-shot}, our generalist \name{} demonstrates better generalization performance than the specialist models. 
Additionally, our model outperforms other LMMs across all tasks in DriveBench, as demonstrated in Tab.~\ref{tab:drivebecnch}.
This validates that the general-purpose all-in-one \name{} can better adapt to new driving scenarios and tasks.
Please refer to the supplementary material for more details.

\subsection{Ablation Study}

In this section, we perform ablation studies to comprehensively validate the effectiveness of our proposed components and report the average scores, including QA enhancement, multi-dataset fine-tuning, curriculum learning, and perspective-aware prompts. Furthermore, we conduct experiments on downstream planning tasks.


\vspace{-5mm}
\paragraph{Question-Answer enhancement.}
Here, we conduct an ablation study on both QA augmentation and QA standardization, with the results presented in Table~\ref{tab:ablation_data}.
Applying diversity augmentation to the datasets leads to substantial performance gains in datasets with constrained question styles, such as CODA-LM and MAPLM (+1.07 and +2.53), as well as in those featuring multiple-choice questions like DriveLM (+12.26).
Furthermore, QA standardization effectively enhances the performance in datasets like DriveLM and NuInstruct (+0.87 and +7.67), which involve detection tasks.
Notably, the performances of all datasets exhibited improvement after applying QA augmentation and QA standardization. 
These results show that such two QA enhancement technologies effectively promote collaboration between multiple datasets.

\vspace{-5mm}
\paragraph{Multi-dataset fine-tuning.}
To validate the mutual enhancement between different datasets, we respectively train the model on the mixed autonomous driving dataset and the individual one. As illustrated in Tab.~\ref{tab:ablation_multidata}, the model trained on the mixed dataset outperforms those trained on the individual dataset. 
This observation underscores the efficacy of mixed training in harnessing the complementary information inherent in diverse datasets, thereby enhancing the model's overall performance and robustness.
Moreover, this approach mitigates the risk of overfitting any single dataset, promoting a more balanced learning process that captures a wider range of features and patterns.

\vspace{-5mm}
\paragraph{Curriculum learning.}
We highlight the critical role of curriculum learning (CL). As shown in~\ref{tab:ablation_training}, \name{} achieves better performance than the baseline without CL which uses mixed 
complexity data from the start (55.71 \vs 61.32). 
As the number of the pre-training steps increases, \name{} achieves progressively improved performance. 
This shows the effectiveness of progressively increasing task and data difficulty in CL.
Moreover, in contrast to traditional CL, we designed a curriculum specifically for AD, which gradually handles unique AD data types and tasks such as multi-view videos and detection.

\vspace{-5mm}
\paragraph{Perspective-aware prompt.}
We assess the effectiveness of perspective-aware prompt (PP) in processing AD data in Tab~\ref{tab:ablation_prompt}.
The comparison shows PP improves performance on multi-view datasets like DriveLM, OmniDrive, and NuInstruct, which involve multi-perspective inputs and related questions. This confirms that including perspectives and data format in instructions helps the model better capture perspective-related features, enhancing its comprehension of multi-view data and spatial perception.

\vspace{-5mm}
\paragraph{Planning performance.}

We finetune our model using nuScenes planning data, and evaluate it on open-loop planning task with L2 error and collision rate,
as shown in Table~\ref{tab:ablation_planning}. Besides strong interactivity and generality, the fine-tuned \name{} outperforms traditional models. 

\label{sec:exp}
\section{Conclusion}
In this paper, we present an all-in-one large multimodal autonomous driving model, \name{}, which can handle various types of data and perform multiple driving tasks in real-world scenarios, demonstrating excellent generality and robustness. 
To our knowledge, we are the first to develop a comprehensive model for AD and evaluate the model across multiple datasets in various AD scenarios.
By augmenting and standardizing several open-source datasets and designing data-related prompts, we conduct multi-step pre-training and fine-tuning of the model from scratch. 
\name{} achieves state-of-the-art performance across data and tasks in the real-world scenarios.
\label{sec:conclusion}
\section{Acknowledgements}
This work is supported by National Key Research and Development Program of China(2024YFE0203100), National Natural Science Foundation of China(NSFC) under Grants No.62476293, Shenzhen Science and Technology Program No.GJHZ20220913142600001, Nansha Key R\&D Program under Grant No.2022ZD014, and General Embodied AI Center of Sun Yat-sen University.

{
    \small
    \bibliographystyle{ieeenat_fullname}
    \bibliography{main}
}

\maketitlesupplementary
\mysection{Implementation Details}
\label{sec:imple}
In this section, we elaborate on the more detailed implementations for \name{} and the experiments in Sec. {\color{red}{3.1}}.
\vspace{-6mm}
\paragraph{Model architecture.}
We adapt SigLIP~\cite{zhai2023sigmoid} as the vision encoder, which is pre-trained on WebLI~\cite{chen2022pali} with a resolution of 384$\times$384.
We use a 2-layer MLP~\cite{liu2024improved} as the projector to project the image features into the word embedding space.
For the language model, we choose Llama-3.1~\cite{meta2024introducing} 8B, which uses a tokenizer with a vocabulary of 128K tokens.
Our model is trained on sequences of 8,192 tokens.
In particular, the vision encoder SigLIP $\mathcal{F}_{\text{e}}(\cdot)$ encodes the input images $X_{\text{v}} \in \mathbb{R}^{(n \times f) \times h \times w \times 3}$ into the visual features:
\begin{equation}
Z_{\text{v}} = \mathcal{F}_{\text{e}}(X_{\text{\text{v}}}),
\end{equation}
where $Z_{\text{v}} \in \mathbb{R}^{(n \times f) \times h' \times w' \times d'}$, $n$ and $f$ denote the number of cameras and frames, $(h, w)$ and $(h', w')$ denote the size of image and feature, and $d'$ denotes the channel dimensionality. 
$X_{\text{v}}$ can represent the data formats mentioned above, for example, $n>1$ and $f>1$ for video from multi-view cameras.
For LiDAR data, we project the point clouds onto the BEV or range view to convert the data into a single image format. 
Afterward, the projector projects MLP $\mathcal{F}_{\text{p}}(\cdot)$ the image features into the word embedding space:
\begin{equation}
H_{\text{v}} = \mathcal{F}_{\text{p}}(Z_{\text{\text{v}}}),
\end{equation}
where the projector is implemented using a 2-layer MLP, $H_{\text{v}} \in \mathbb{R}^{(n \times f \times h \times w) \times d}$ denotes a sequence of visual tokens, and $d$ denotes the dimensionality of the word embedding space in LLM LlamA-3.1 $\mathcal{F}_{\text{l}}(\cdot)$.
Based on the visual tokens $H_{\text{v}}$ and the user instruction $X_{\text{t}}$, the LLM computes the probability of the target word step by step:
\begin{equation}
    p( Y_{\text{t}} |  H_{\text{v}}, X_{\text{t}}) =
    \prod_{i=1}^{L} \mathcal{F}_{\text{l}} (  Y_{\text{t}, i}
| H_{\text{v}}, \mathrm{\Phi}(X_{\text{t}}), Y_{\text{t}, 0:i-1}),
    \label{eq:llm}
\end{equation}
where $\mathrm{\Phi}(\cdot)$ refers the text tokenizer, $Y_{\text{t}, i}$ and $Y_{\text{t}, 0:i-1}$ represent the $i$th word and the preceding $i-1$ words in $Y_{\text{t}}$, and $L$ indicates the length of the words generated by LLM.


\paragraph{Dataset.}
This part further introduces the evaluation datasets applied in Sec. {\color{red}{4.1}}, including six datasets for general capability, three datasets for testing generalization abilities, and one dataset designed for planning tasks.
\begin{itemize}
 \item DriveLM, OmniDrive and Nuinstruct are annotated based on the nuScenes\cite{caesar2020nuscenes} dataset and contain 376,181, 374,329 and 71,842 samples respectively, suitable for multi-view images or videos input.
The CODA-LM dataset, designed for corner case question-answer pairs on the CODA \cite{li2024automated} dataset, includes a total of 184,480 samples in both Chinese and English, as a single-view dataset.
MAPLM, a multi-view image dataset, contains 94,970 samples, while LingoQA, a single-view video dataset, comprises 413,829 samples.
Note the sample number is computed after our enhancement.
Additionally, each dataset includes a specific number of test samples: DriveLM has 15,480 test samples, OmniDrive has 72,184 test samples, NuInstruct has 16,147 test samples, CODA-LM has 2,123 test samples, MAPLM has 6,642 test samples, and LingoQA has 500 test samples.
 \item In zero-shot setting, we use BDD-X, DRAMA and DriveBench dataset.
BDD-X is composed of over 77 hours of driving within 6,970 videos. The test set of BDD-X consists of 698 driving videos. The videos are taken in diverse driving conditions, \eg~day/night, highway/city/countryside, summer/winter \etc~On average 40 seconds long, each video contains around 3-4 actions, \eg~speeding up, slowing down, turning right etc., all of which are annotated with a description and an explanation.
DRAMA dataset is captured from a moving vehicle on highly interactive urban traffic scenes in Tokyo.
It contains different annotations: Video-level Q/A, Object-level Q/A, Risk object bounding box, Free-form caption, and separate labels for ego-car intention, scene classifier and suggestions to the driver.
The DRAMA test set comprises 2544 samples.
DriveBench is a benchmark dataset designed to assess VLM reliability across 17 settings, including clean, corrupted, and text-only inputs. It comprises 19,200 frames, 20,498 question-answer pairs, three question types, and four mainstream driving tasks, evaluating 12 popular VLMs.
 \item We evaluate our model in the clean setting and compare it with all models in the dataset, which covers four mainstream driving tasks: perception, prediction, planning, and behavior.
We utilize GPT-4 for GPT Score evaluation in BDD-X and DRAMA, while employing GPT-3.5-turbo for DriveBench. 
We use nuScenes to evaluate the planning task, which is a challenging and popular benchmark in the AD. The dataset is a multi-sensor dataset
with 1,000 scenes and each scene lasts for 20 seconds. There are 6,019 validation samples.
\end{itemize}

\paragraph{Experiment setting.}
Here we describe the training details of \name{}. We adopt a curriculum learning approach to progressively train \name{} as introduced in Sec. {\color{red}{3.4}}.
\begin{itemize}
    \vspace{2mm}
    \item \emph{Stage-1: Language-image alignment.} 
    We use LCS-558K~\cite{liu2024visual} to align the visual patch features into the word embedding space. 
    During this stage, we train only the projector while keeping the other components frozen. 
    The learning rate is set to 1$\times 10^{-3}$ and the training is conducted for 1 epoch with a batch size of 512.

    \vspace{2mm}
    \item \emph{Stage-2: Single-image pre-training.}
    In this stage, we use single-image data to improve model's image comprehension capability. 
    We utilize the recaptioned BLIP558K~\cite{liu2024visual}, COCO118K~\cite{li2024llava} and CC3M~\cite{sharma2018conceptual} datasets to improve the model. 
    Meanwhile, we use language data Evo-Instruct~\cite{chen2024allava} to balance the model's language understanding ability.
    At this stage, the dataset comprises 3M single-image data and 143K language data.
    We fine-tune the entire model using a batch size of 256 for 1 epoch. 
    The learning rate for the vision encoder is set to 2$\times 10^{-6}$, while the learning rate for both the projector and LLM is 1$\times 10^{-5}$.

    \vspace{2mm}
    \item \emph{Stage-3: Multi-capacity pre-training.}
    In this stage, our primary objective is to enhance the model's reasoning and perception capabilities, and equip the model with the ability to handle diverse data formats. 
    To achieve this, we use various multimodal data and perception data, including 1.5M single images, 760K multi-view images, 501K single videos, and 145K multi-view videos.
    Specifically, the single-image data consists of the multimodal data from~\cite{li2024llava} and the perception data from COCO~\cite{lin2014microsoft} and Object365~\cite{shao2019objects365}. 
    The multi-view image data includes the multimodal data LLaVA-NeXT-Interleave~\cite{li2024llavainterleave} and the perception data nuScenes~\cite{caesar2020nuscenes}. The single-view video data is derived from the works~\cite{li2024llavainterleave,zhang2024video}.
    Given the scarcity of multi-view videos in the available data, we generate the multi-view perception data using nuScenes~\cite{caesar2020nuscenes}, with each view consisting of 5 frames. 
    We fine-tune the entire model with a batch size of 256 for 1 epoch, maintaining the same learning rates as Stage-2.

    \vspace{2mm}
    \item \emph{Stage-4: Driving fine-tuning.}
    In the final stage, we employ a diverse array of high-quality driving datasets to fine-tune \name{} for AD tasks. 
    We compile six public AD datasets, which include single image (CODA-LM), multi-view images (MAPLM, DriveLM), single video (LingoQA), and multi-view videos (OmniDrive, NuInstruct), amounting to a total of 1.5M.
    Specifically, we augment the CODA-LM dataset, expanding it from 36,896 to 184,480 samples, and the MAPLM dataset from 47,485 to 94,970 samples. Additionally, we standardize the DriveLM and NuInstruct datasets to ensure uniformity across the data.
    In this stage, we use the same batch size and learning rate as Stage-2.
\end{itemize}

\mysection{Prompt Design}
{
\begin{table}[H]
\begin{tcolorbox}[colback=gray!10,
                  colframe=black,
                  width=\linewidth,
                  arc=1mm, auto outer arc,
                  boxrule=0.5pt,
                 ]
{
\small
\textcolor{Blue}{\textbf{Messages}}
 = [ 

\smallskip
 \{\textbf{\texttt{"}role\texttt{"}: \texttt{"}system\texttt{"},
\texttt{"}content\texttt{"}:} f\texttt{"""}You are an English improver.\texttt{"""}\},

\smallskip
\{\textbf{\texttt{"}role\texttt{"}: \texttt{"}\textbf{user}\texttt{"},
\texttt{"}content\texttt{"}:} f\texttt{"""}I have a question and its corresponding answer. I need your assistance in revising and refining them. Please make some changes to the written content while preserving the meaning. The question and answer that require modifications are:
\{\textcolor{Blue}{QA}\}. Please provide the revised question and answer in the format: Question: \texttt{<}question\texttt{>} Answer: \texttt{<}answer\texttt{>}.\texttt{"""}\}]}

\end{tcolorbox}
\vspace{-2mm}
\caption{The prompt for question-answer augmentation.}
\label{tab:a-prompt}
\end{table}
}
In the question-answer augmentation, we use the prompt presented in Tab.~\ref{tab:a-prompt} for GPT-4o-mini to enhance and improve the question-answer pairs.

\begin{figure*}[h]
  \centering
  \includegraphics[width=\linewidth]{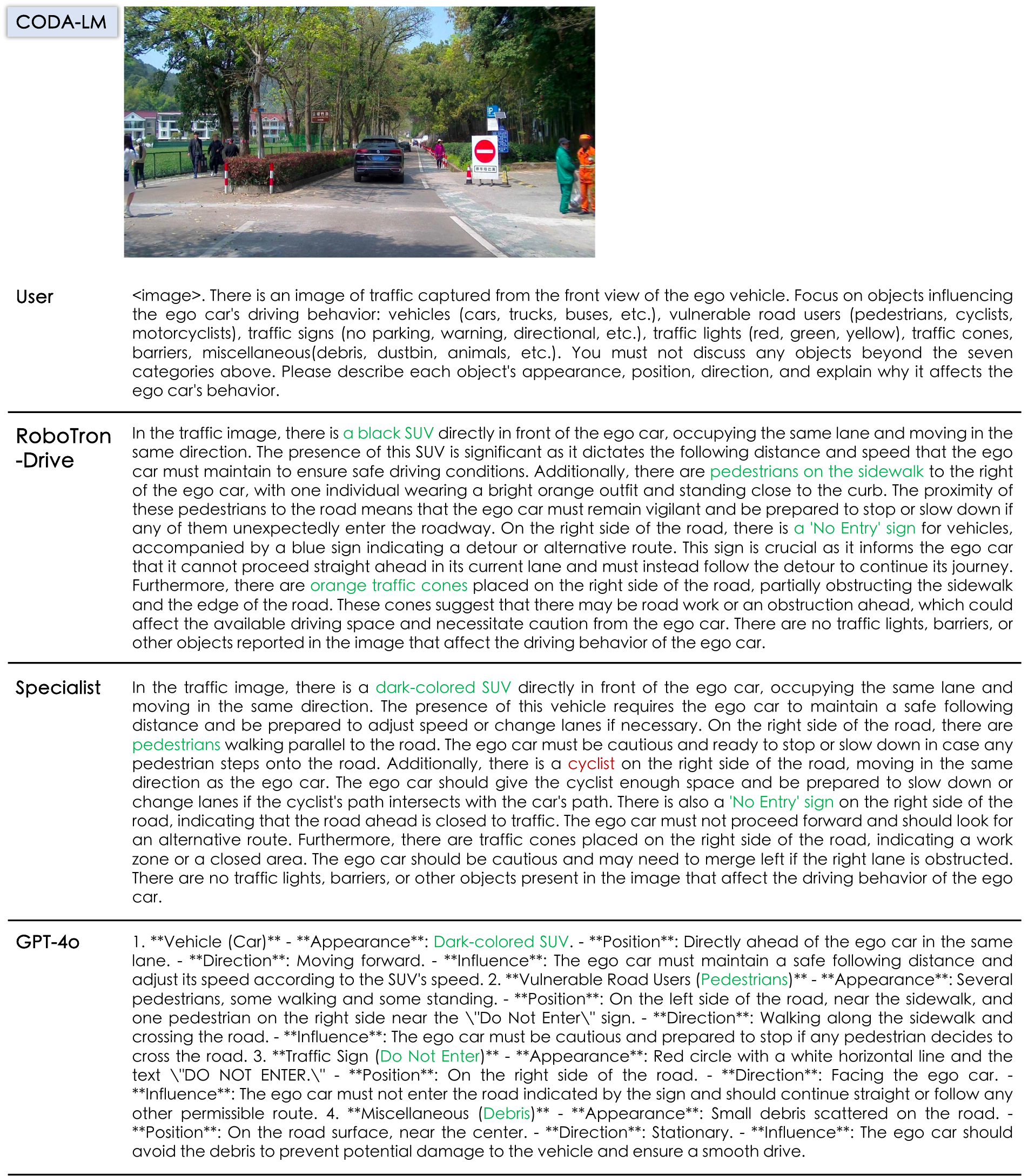}
  \caption{Visualization of CODA-LM. 
  Key information is highlighted in {\color{green}{green}}, while errors are marked in {\color{red}{red}}.}
  \label{fig:appendixf}
\end{figure*}
\begin{figure*}[h]
  \includegraphics[width=\linewidth]{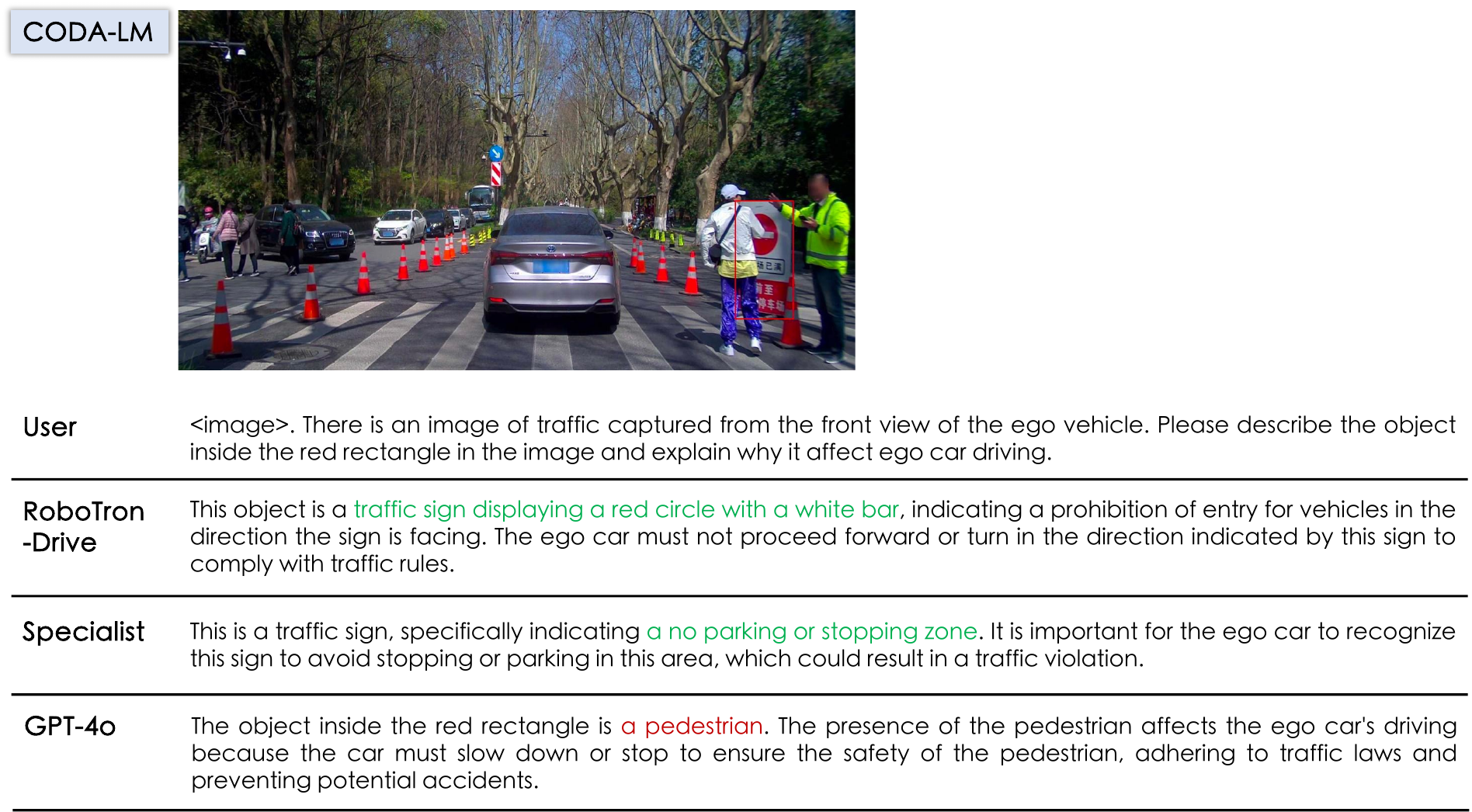}
  \caption{Visualization of CODA-LM.}
\label{fig:appendix1}
\end{figure*}
\begin{figure*}[h]
  \includegraphics[width=\linewidth]{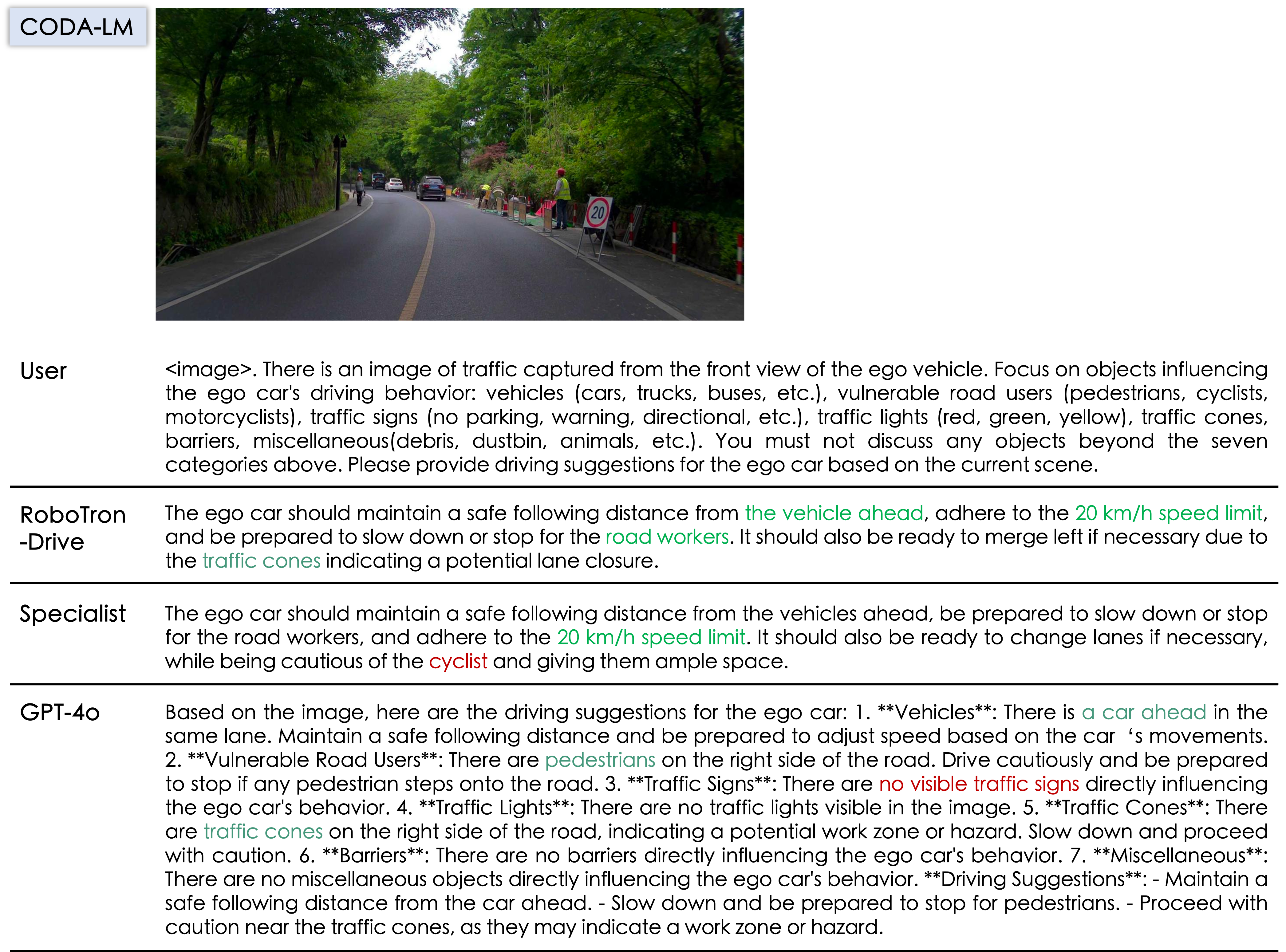}
  \caption{Visualization of CODA-LM.}
\label{fig:appendix2}
\end{figure*}
\begin{figure*}[h]
  \includegraphics[width=\linewidth]{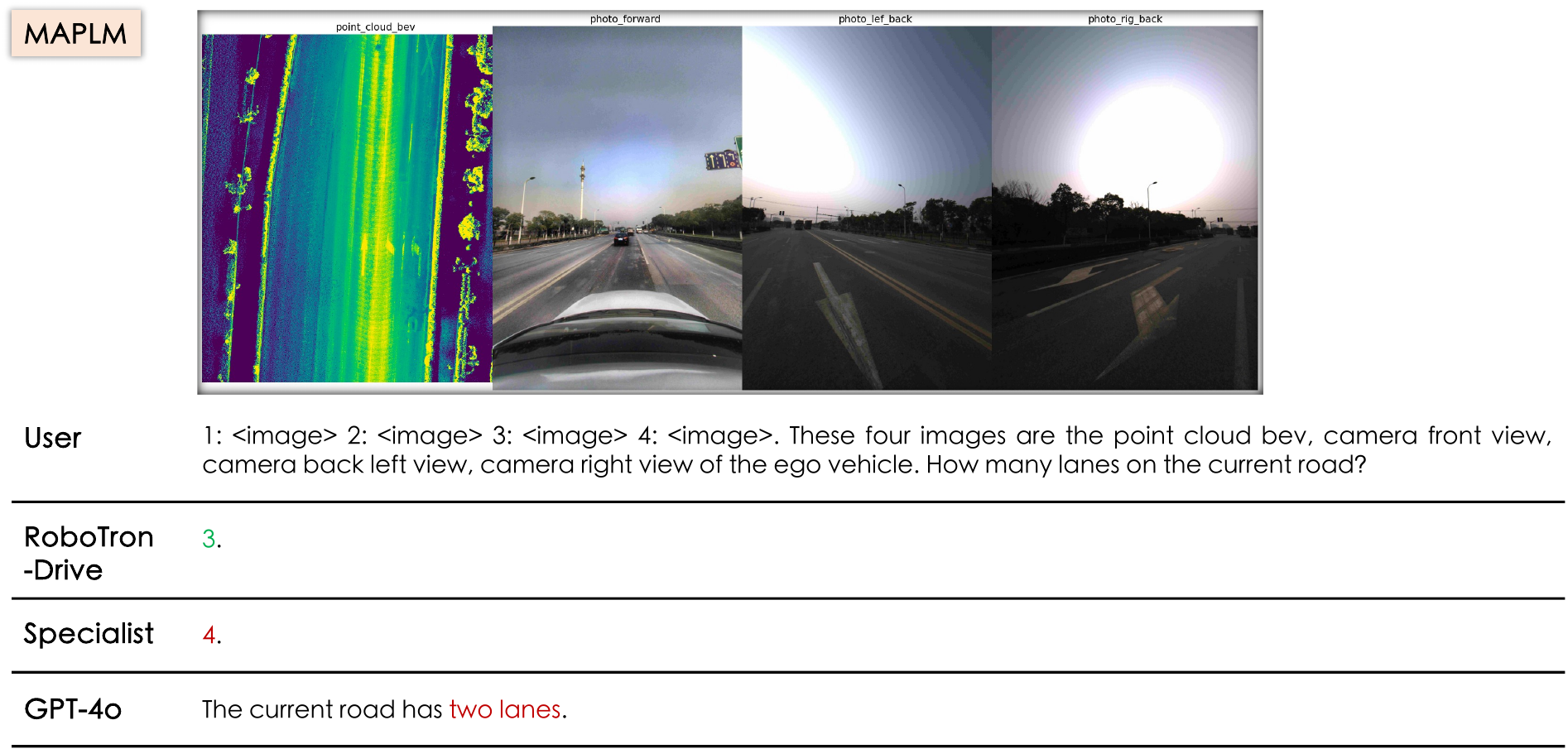}
  \caption{Visualization of MAPLM.}
\label{fig:appendix3}
\end{figure*}
\begin{figure*}[h]
  \includegraphics[width=\linewidth]{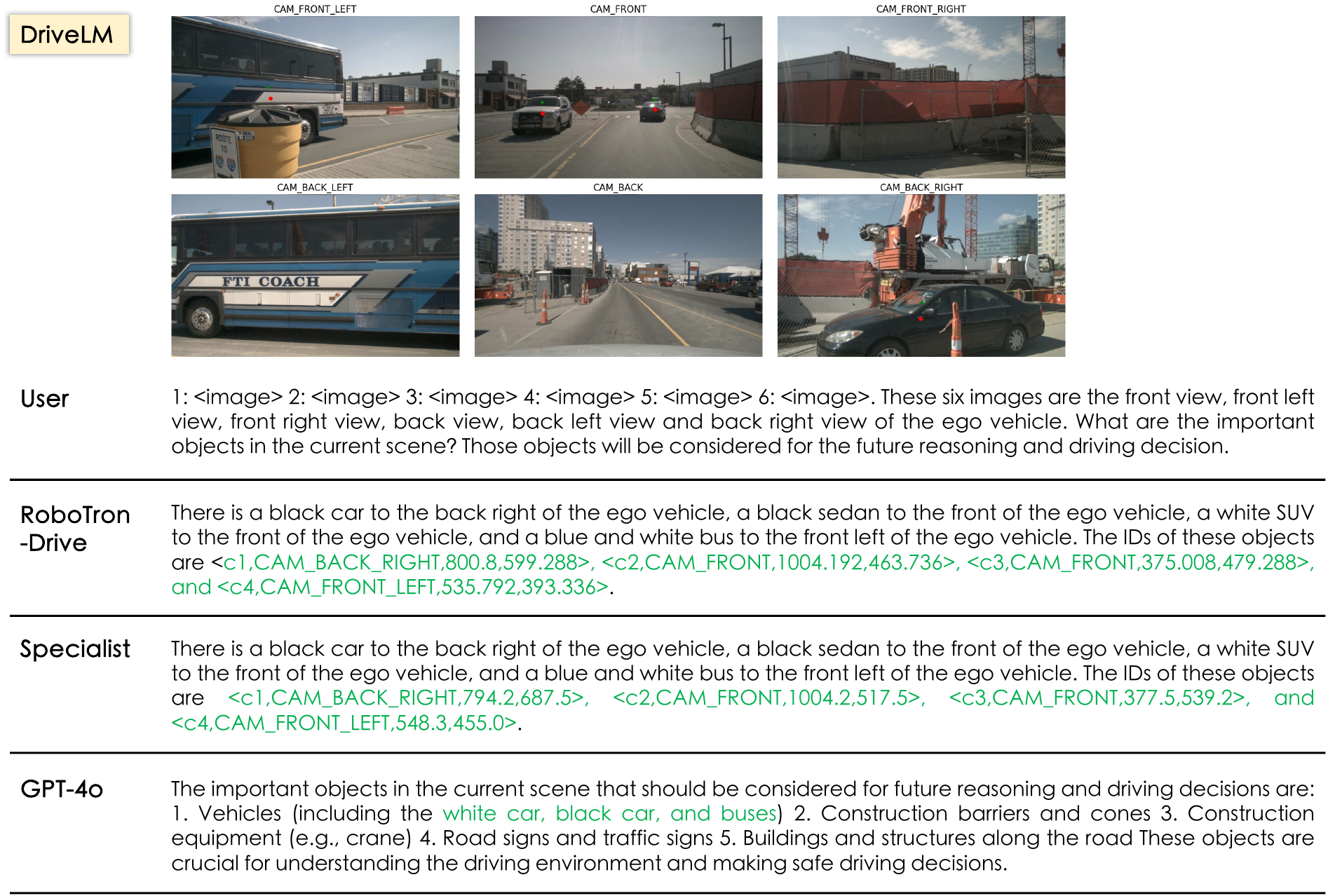}
  \caption{Visualization of DriveLM.}
\label{fig:appendix4}
\end{figure*}
\begin{figure*}[h]
  \includegraphics[width=\linewidth]{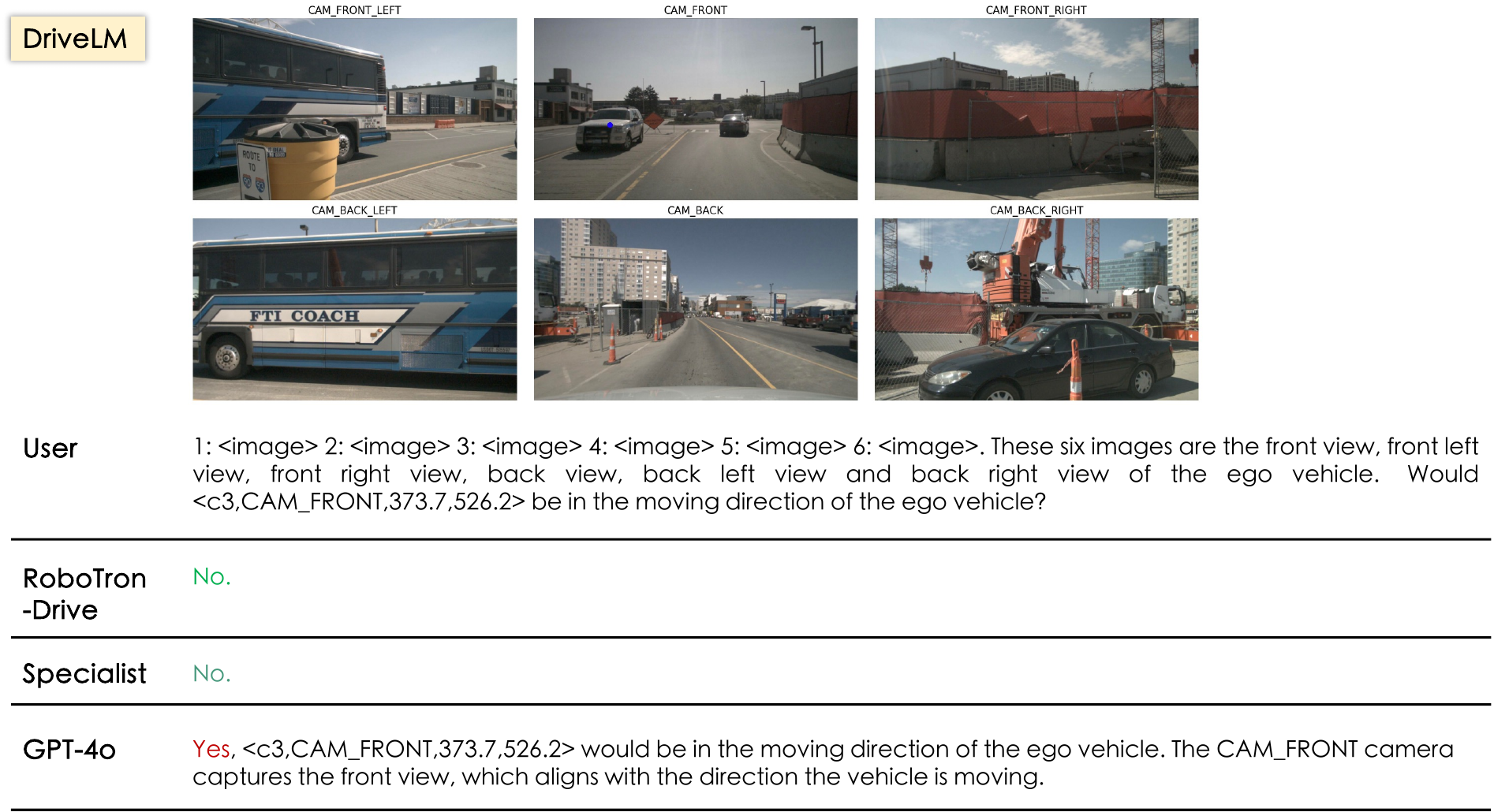}
  \caption{Visualization of DriveLM.}
\label{fig:appendix5}
\end{figure*}
\begin{figure*}[h]
  \includegraphics[width=\linewidth]{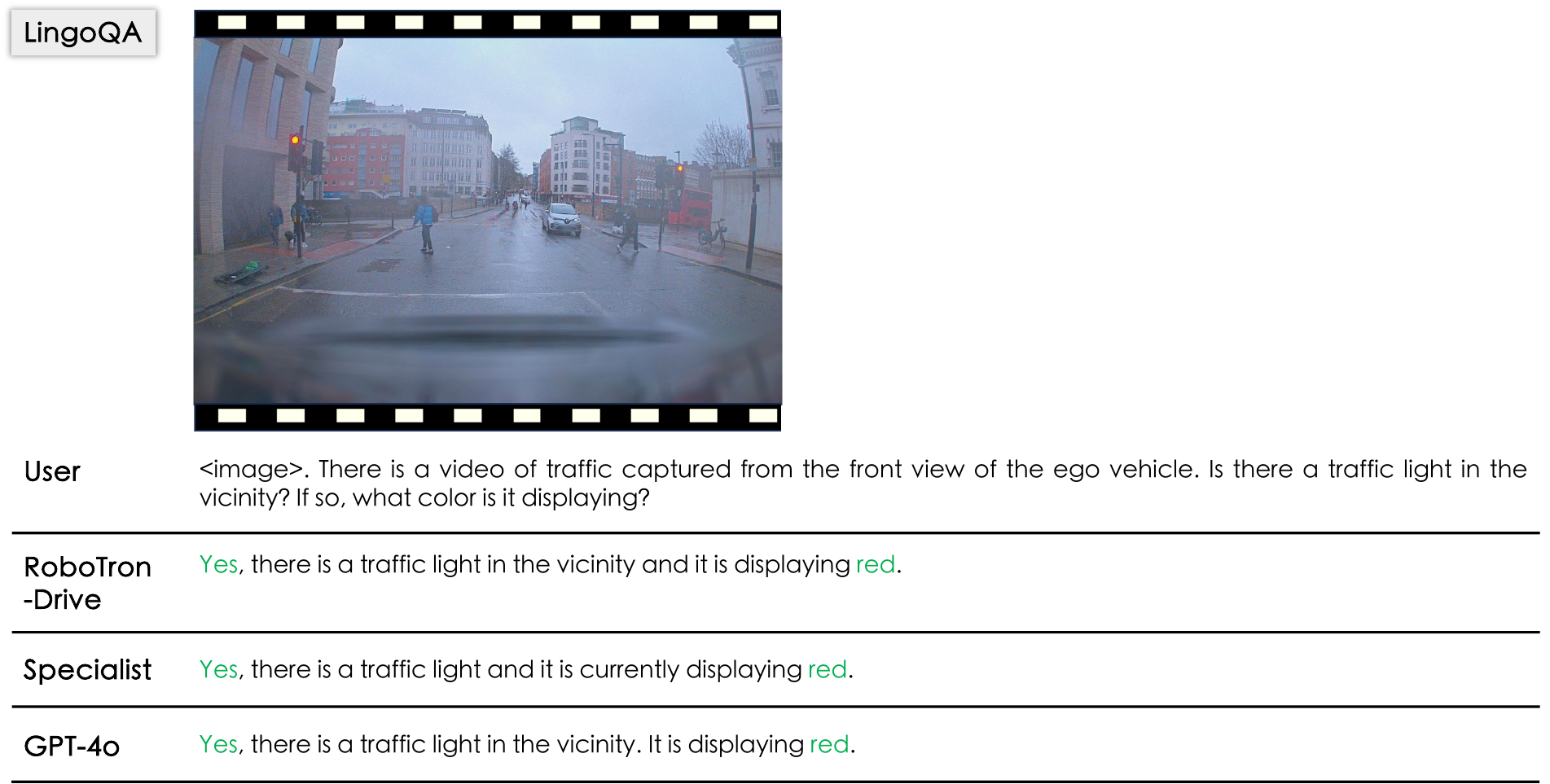}
  \caption{Visualization of LingoQA.}
\label{fig:appendix6}
\end{figure*}
\begin{figure*}[h]
  \includegraphics[width=\linewidth]{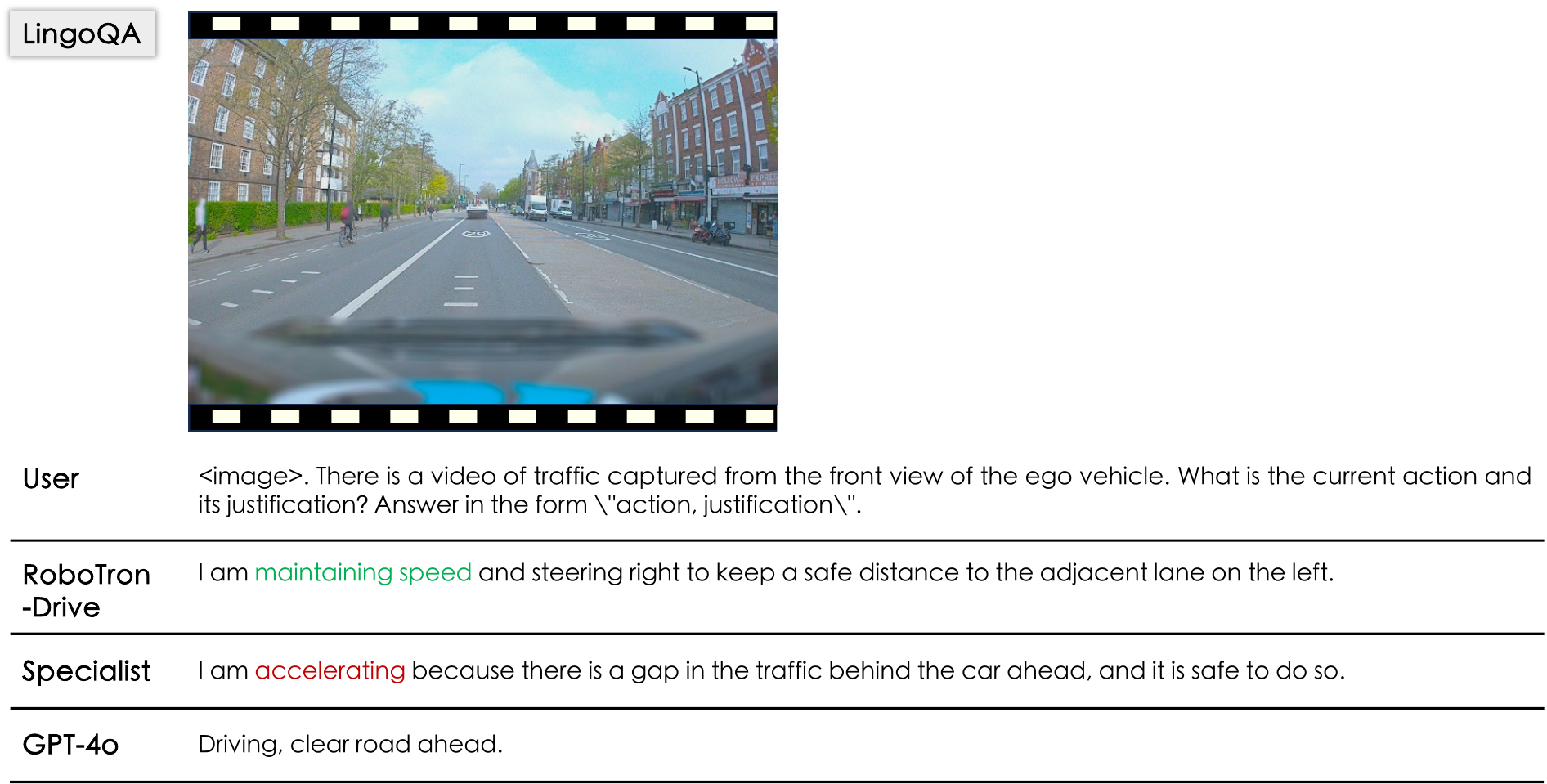}
  \caption{Visualization of LingoQA.}
\label{fig:appendix7}
\end{figure*}
\begin{figure*}[h]
  \includegraphics[width=\linewidth]{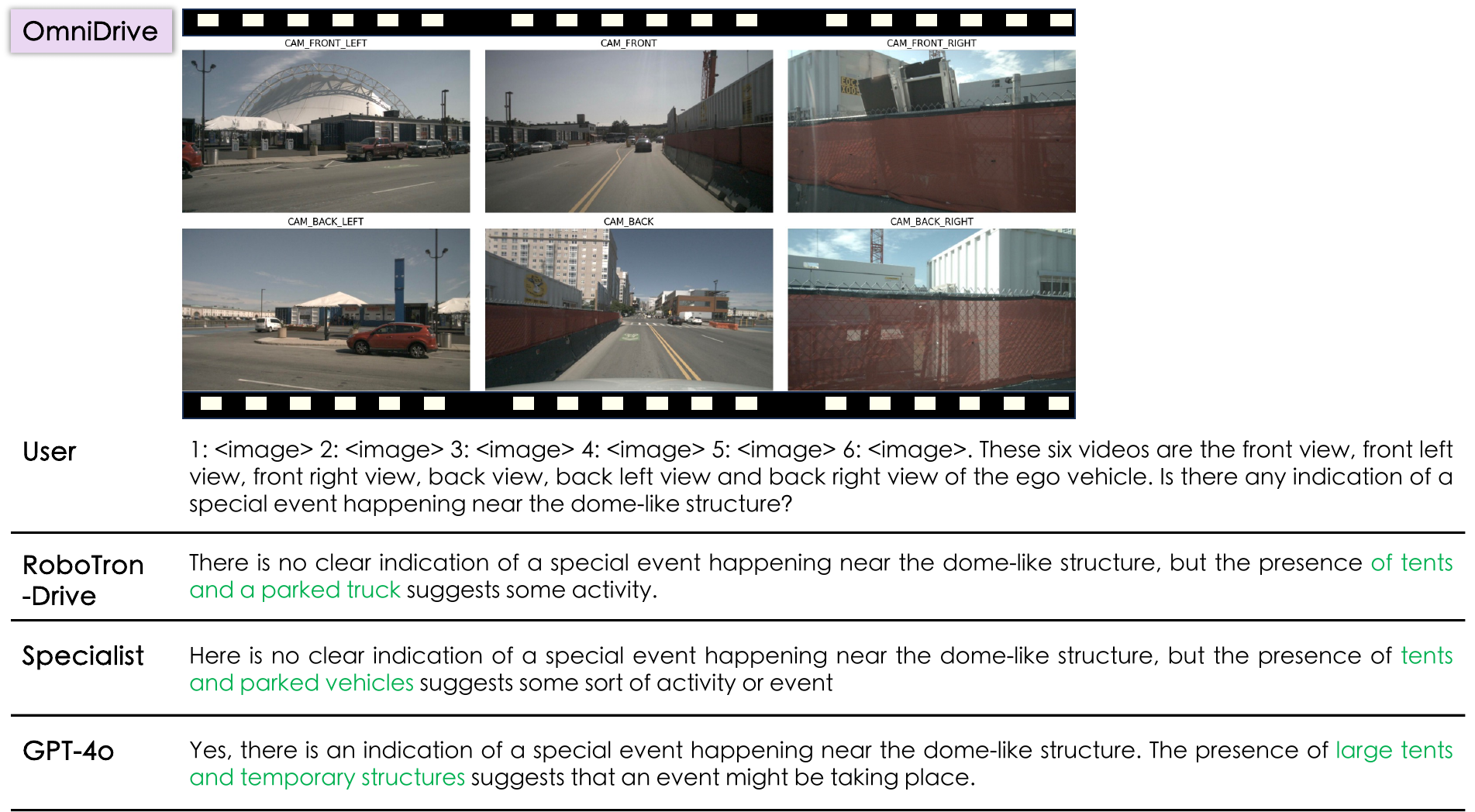}
  \caption{Visualization of OmniDrive.}
\label{fig:appendix8}
\end{figure*}
\begin{figure*}[h]
  \includegraphics[width=\linewidth]{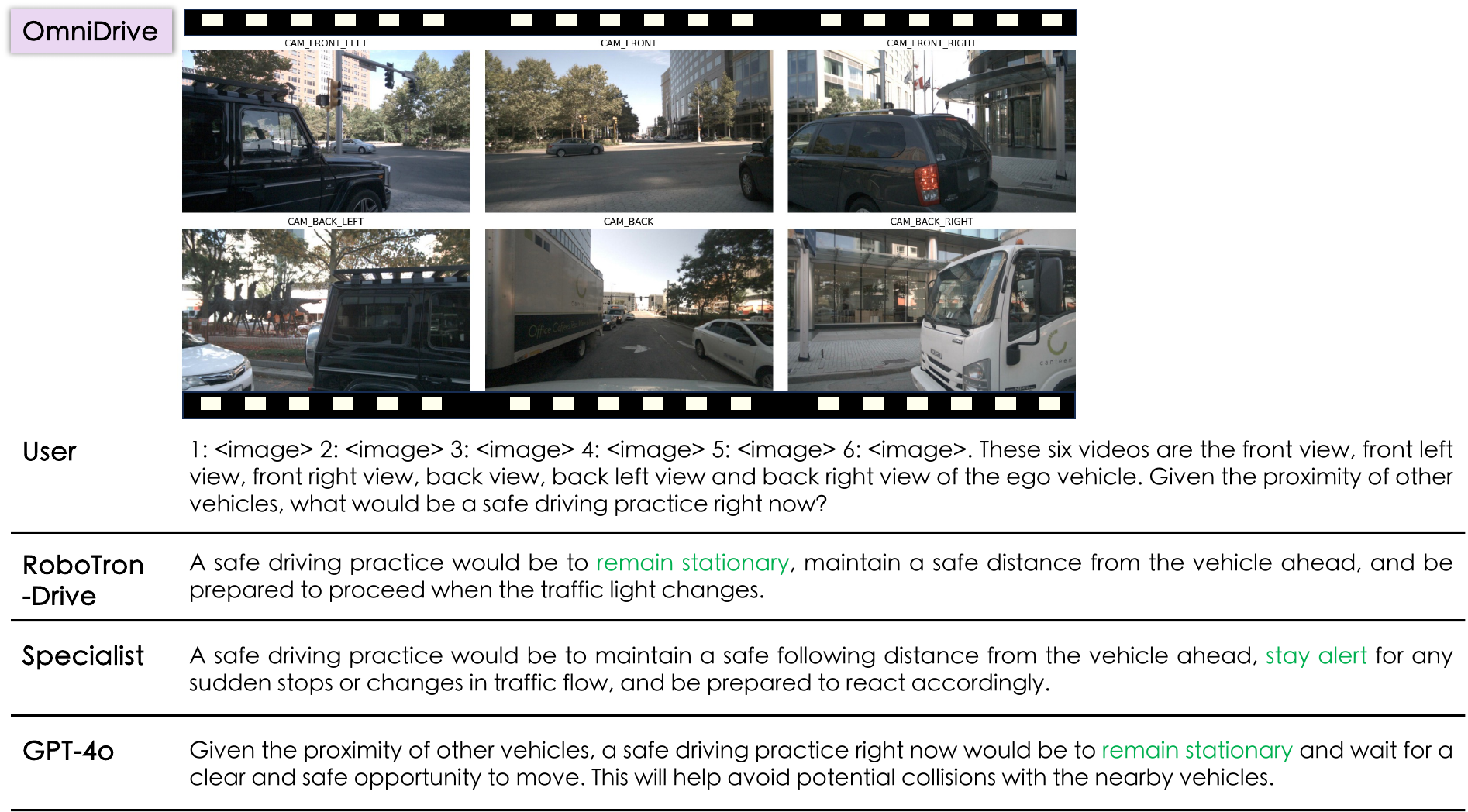}
  \caption{Visualization of OmniDrive.}
\label{fig:appendix9}
\end{figure*}
\begin{figure*}[h]
  \includegraphics[width=\linewidth]{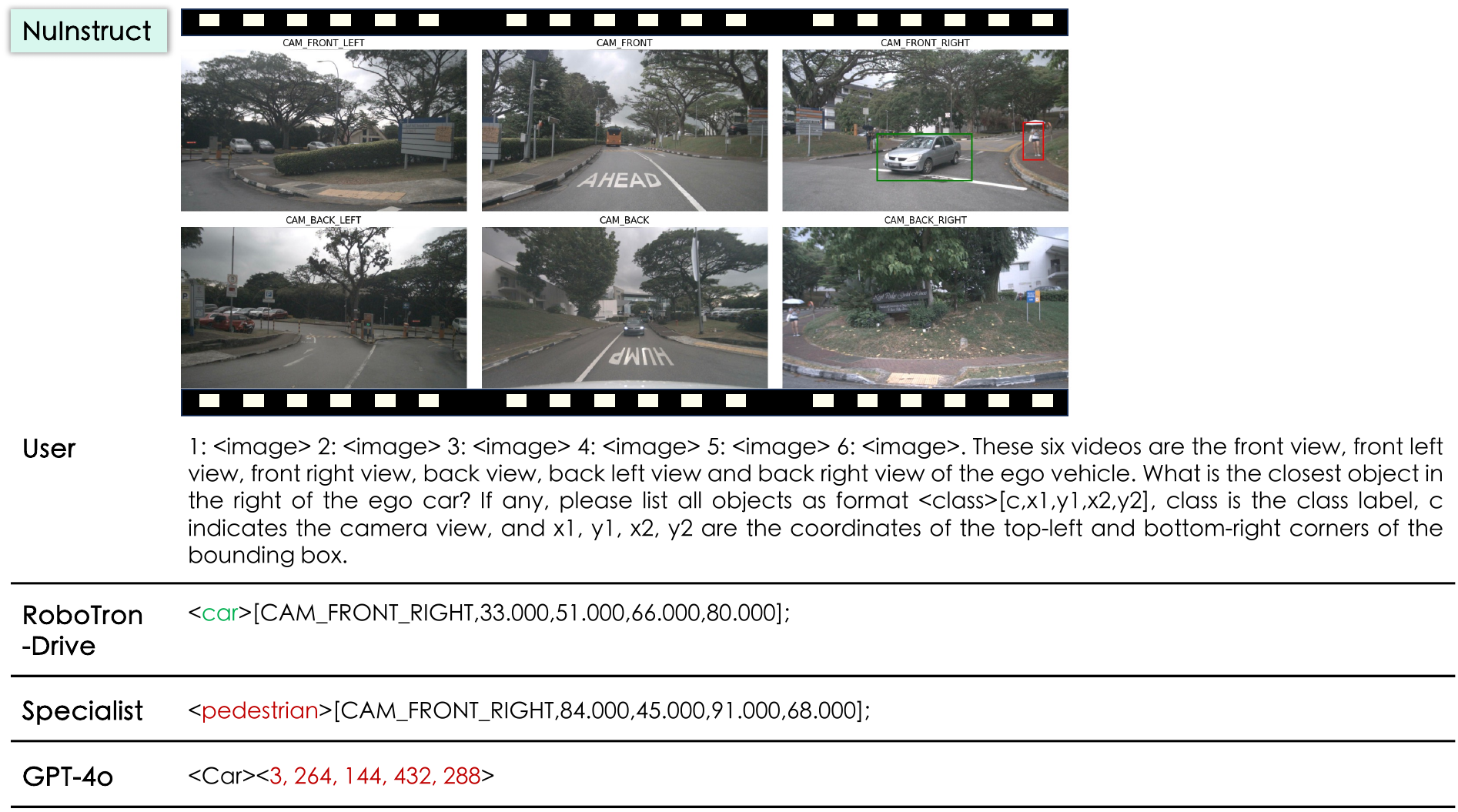}
\caption{Visualization of NuInstruct.}
\label{fig:appendix10}
\end{figure*}
\begin{figure*}[h]
  \includegraphics[width=\linewidth]{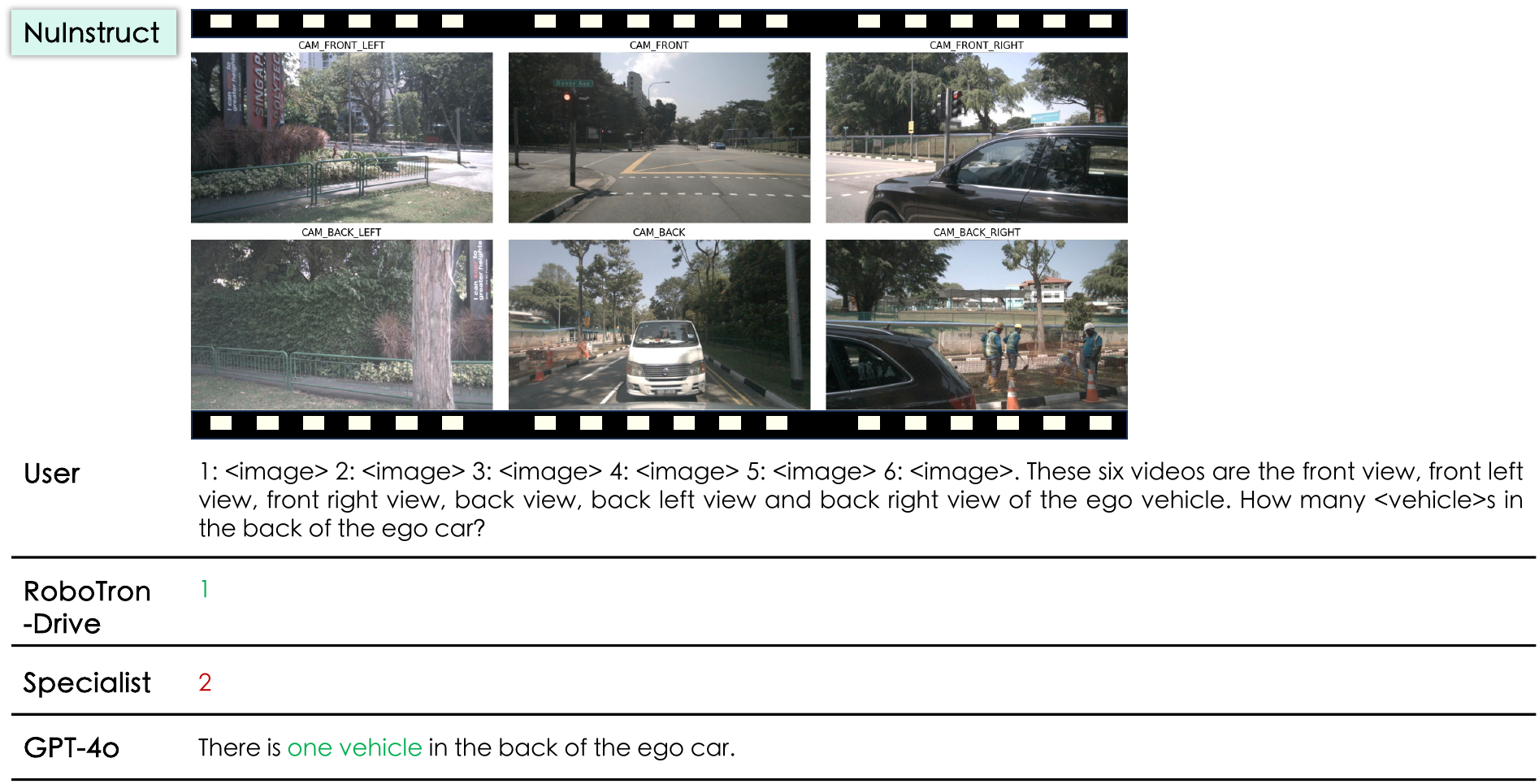}
\caption{Visualization of NuInstruct.}
\label{fig:appendix11}
\end{figure*}
\begin{figure*}[h]
  \includegraphics[width=\linewidth]{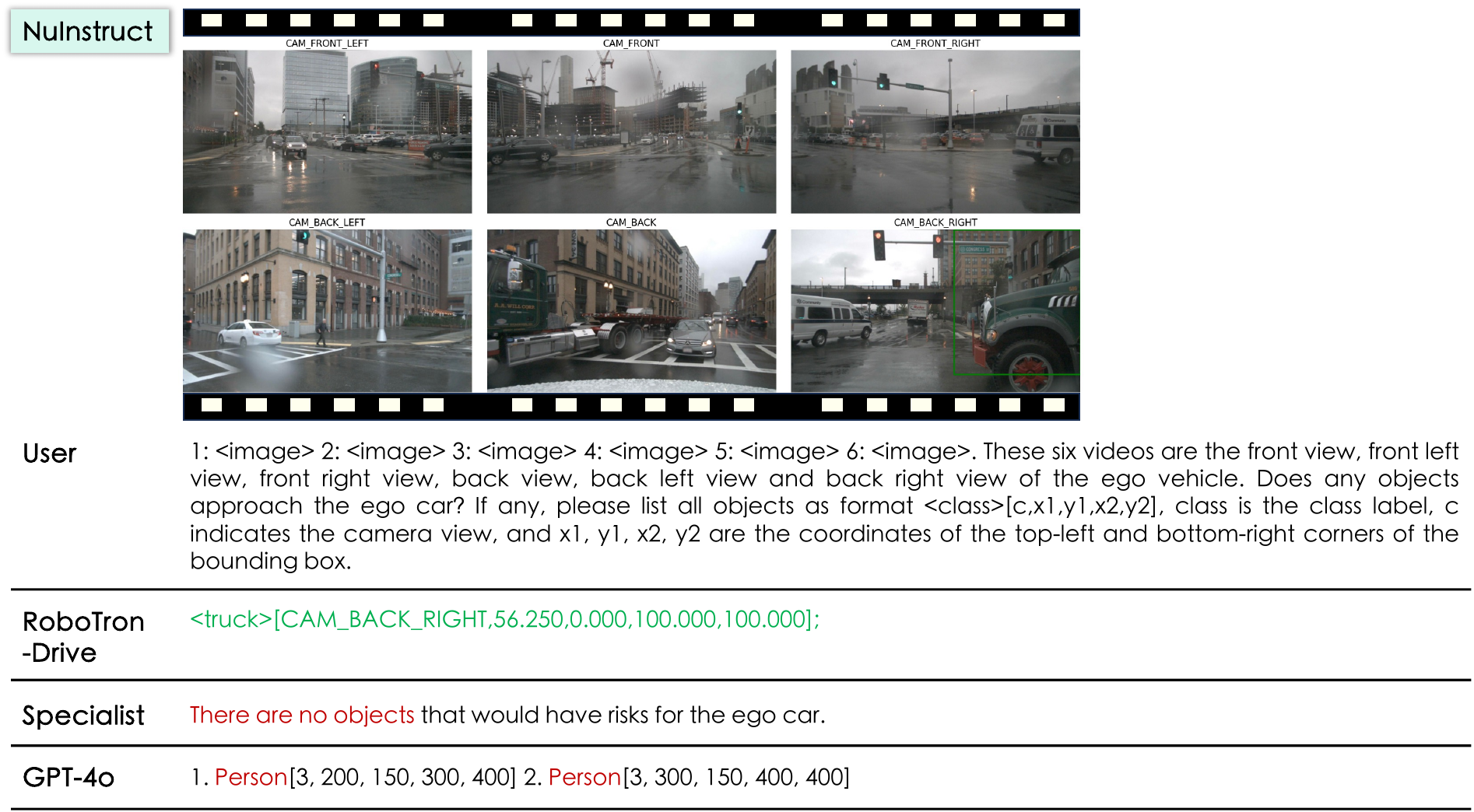}
\caption{Visualization of NuInstruct.}
\label{fig:appendix12}
\end{figure*}
\begin{figure*}[h]
  \includegraphics[width=\linewidth]{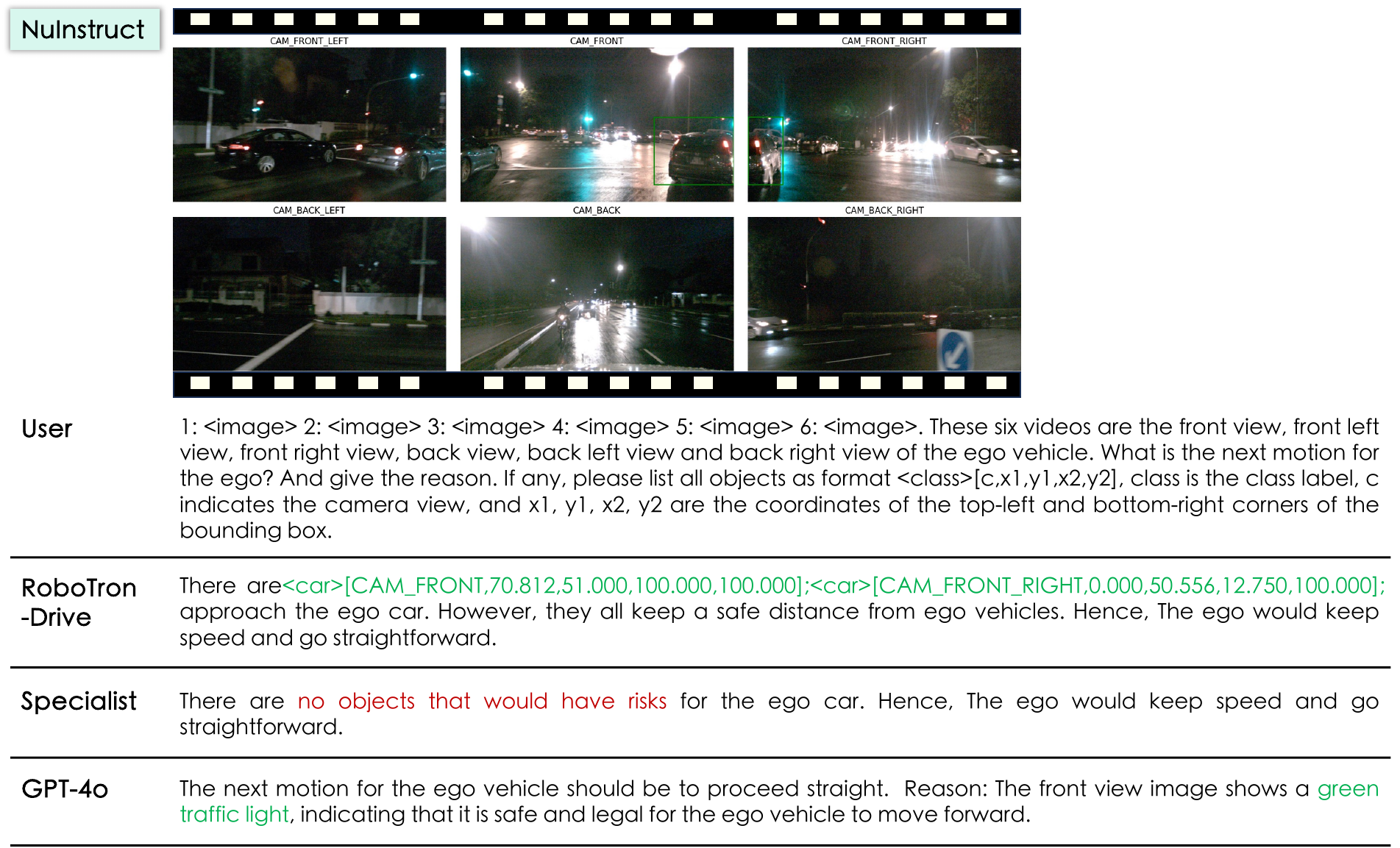}
\caption{Visualization of NuInstruct.}
\label{fig:appendix13}
\end{figure*}
\begin{figure*}[h]
  \includegraphics[width=\linewidth]{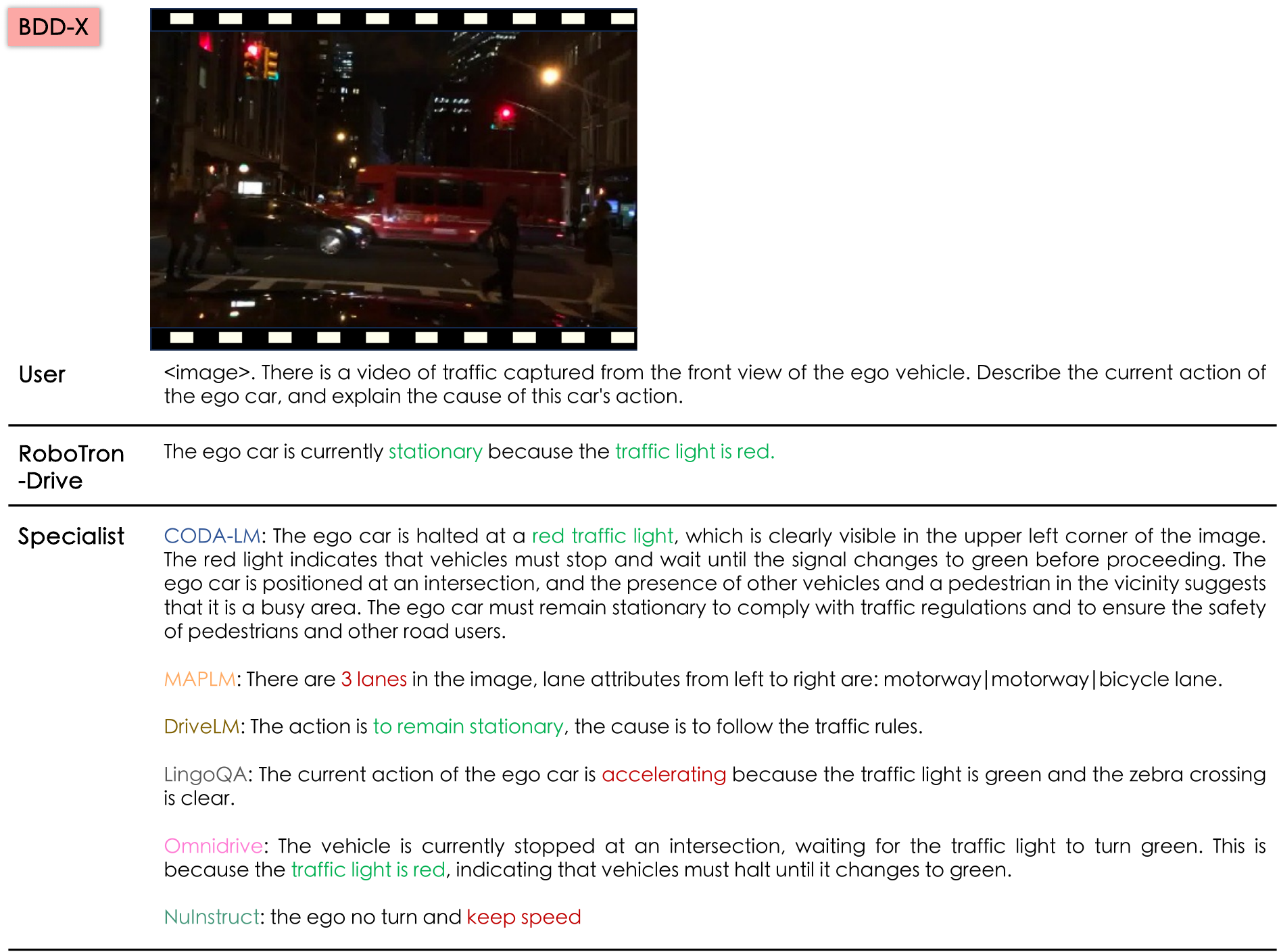}
  \caption{Visualization of BDD-X.}
\label{fig:appendix14}
\end{figure*}
\begin{figure*}[h]
  \includegraphics[width=\linewidth]{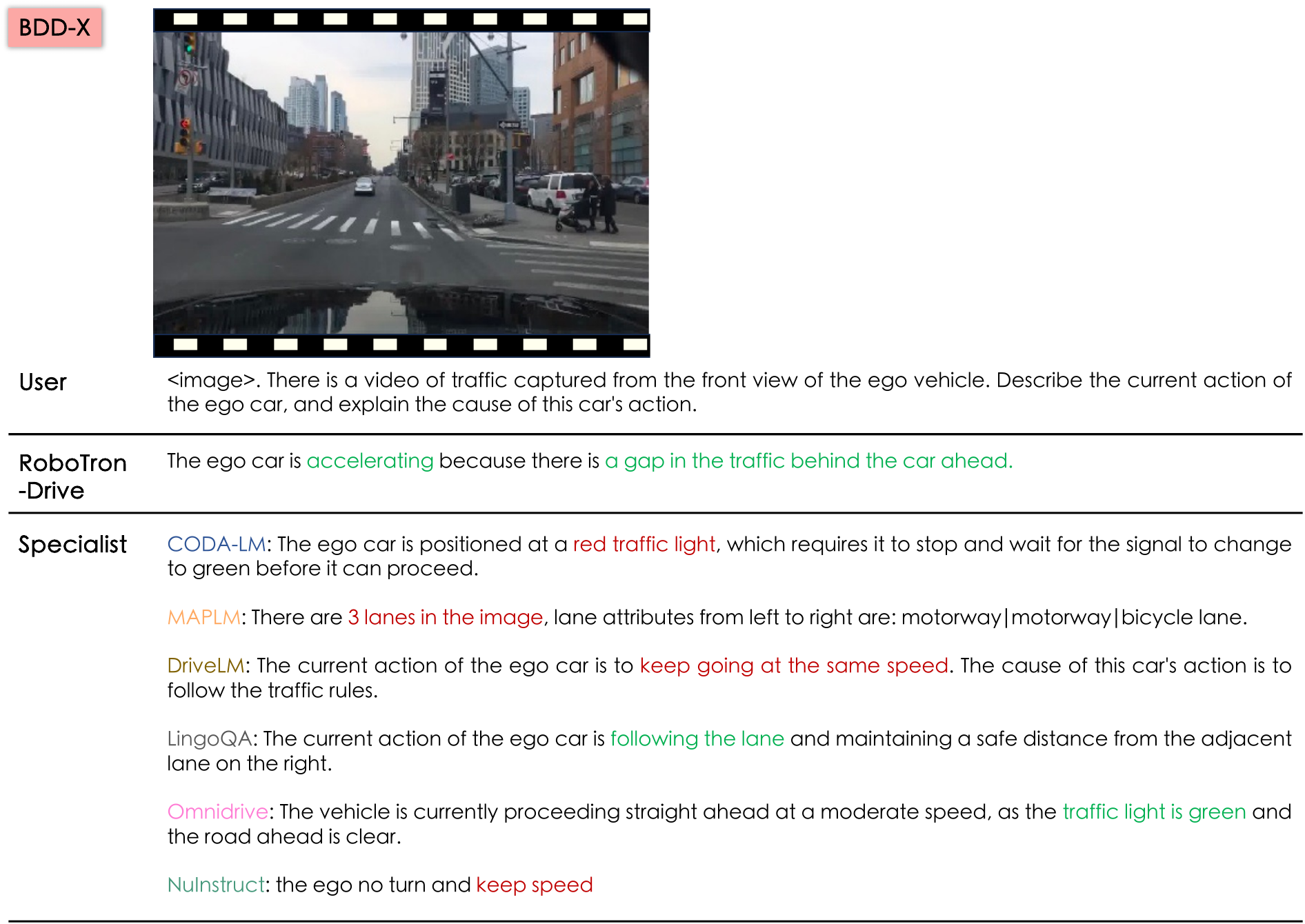}
  \caption{Visualization of BDD-X.}
\label{fig:appendixl}
\end{figure*}

\mysection{Qualitative Results}
Here we present the qualitative results for analyzing the distinctions among various models.
In Fig.~\ref{fig:appendixf}-\ref{fig:appendixl}, we provide the visualization examples from each dataset, illustrating the general performance on the six benchmarks and the generalization ability in BDD-X.
The specialist model is trained on individual dataset. 
The all-in-one model \name{} is able to accurately identify and analyze key objects related to driving in images and videos, thereby enhancing its ability to handle diverse autonomous driving data and tasks.

\onecolumn

\twocolumn


\end{document}